\documentclass[journal]{IEEEtran}
\usepackage{xcolor}
\ifCLASSINFOpdf
\usepackage{subfig}
  \usepackage[pdftex]{graphicx}
  \graphicspath{{./images/}}
  \DeclareGraphicsExtensions{.pdf,.jpeg,.png}
\else
\fi
\usepackage{tabularx,booktabs,multicol,multirow,makecell}
\usepackage{amsmath}
\usepackage{amssymb}
\usepackage{xfrac}
\renewcommand{\vec}[1]{\boldsymbol{#1}}

\usepackage[ruled,vlined,boxed]{algorithm2e}
\usepackage{wrapfig}
\usepackage{url}

\usepackage{color}

\definecolor{RoyalBlue}{cmyk}{1, 0.50, 0, 0}

\usepackage{tikz}
\usepackage{textcomp}
\usepackage{hyperref}
\usepackage{lipsum}

\newcommand\copyrighttext{%
  \footnotesize \textcopyright 2022 IEEE. Personal use of this material is permitted.  Permission from IEEE must be obtained for all other uses, in any current or future media, including reprinting/republishing this material for advertising or promotional purposes, creating new collective works, for resale or redistribution to servers or lists, or reuse of any copyrighted component of this work in other works.
  DOI: \href{https://ieeexplore.ieee.org/document/9780217}{10.1109/TKDE.2022.3177368}}
\newcommand\copyrightnotice{%
\begin{tikzpicture}[remember picture,overlay]
\node[anchor=south,yshift=10pt] at (current page.south) {\fbox{\parbox{\dimexpr\textwidth-\fboxsep-\fboxrule\relax}{\copyrighttext}}};
\end{tikzpicture}%
}

\begin{document}
%
\title{LAAT: Locally Aligned Ant Technique for discovering multiple faint low dimensional structures of varying density}
%
%
%

\author{Abolfazl~Taghribi,~
	Kerstin~Bunte,~
	Rory~Smith,~
	Jihye Shin,~
	Michele Mastropietro,~
	Reynier F. Peletier,~
	and Peter~Ti\v{n}o
\thanks{A. Taghribi and K. Bunte are with the 
Faculty of science and engineering, University of Groningen, NL 
e-mail: abolfazl.taghribi@gmail.com.}
\thanks{R. Smith is with Universidad Técnica Federico Santa María}
\thanks{J. Shin is with Korea Astronomy and Space Science Institute.}
\thanks{M. Mastropietro is with Ghent University.}
\thanks{R. F. Peletier is with University of Groningen.}
\thanks{P. Tino is with the University of Birmingham.}
}

%
%

\markboth{IEEE Transactions on Knowledge and Data Engineering,~Vol.~x, No.~x, Month~2022}%
{Taghribi \MakeLowercase{\textit{et al.}}: Bare Demo of IEEEtran.cls for IEEE Journals}
%



\maketitle
\copyrightnotice
\begin{abstract}
Dimensionality reduction and clustering are often used as preliminary steps for many complex machine learning tasks.  
The presence of noise and outliers can deteriorate the performance of such preprocessing and therefore impair the subsequent analysis tremendously. 
In manifold learning, several studies indicate solutions for removing background noise or noise close to the 
structure when the density is substantially higher than that exhibited by the noise. 
However, in many applications, including astronomical datasets, the density varies alongside manifolds that are buried in a noisy background.
We propose a novel method to extract manifolds in the presence of noise based on the idea of Ant colony optimization. In contrast to the existing random walk solutions, our technique captures points that are locally aligned with major directions of the manifold. 
Moreover, we empirically show that the biologically inspired formulation of ant pheromone reinforces this behavior enabling it to recover multiple manifolds embedded in extremely noisy data clouds.
The algorithm performance in comparison to state-of-the-art approaches for noise reduction in manifold detection and clustering is demonstrated, on several synthetic and real datasets, including an N-body simulation of a cosmological volume.
\end{abstract}


%
\IEEEpeerreviewmaketitle

\section{Introduction}
\IEEEPARstart{M}ANY fields such as astronomy \cite{libeskind_tracing_2018}, genetics \cite{sardiu_identification_2017,camara_topological_2017}, 
medical science \cite{torres_tracking_2016,nielson_uncovering_2017}, and sensory activity recognition \cite{chazal_subsampling_2015} deal with a large number of high-dimensional data points located on some lower-dimensional structure hidden inside big data point clouds. 
Nonlinear dimensionality reduction methods \cite{roweis_nonlinear_2000} and manifold learning techniques \cite{wu_structure-aware_2018} try to find the underlying low dimensional manifold and project the data on a lower-dimensional space to save computation in the next steps of analysing the data. 
However, in real-world applications, sample points may exist on several 
linear or nonlinear manifolds surrounded by noise and outliers which depreciate the result of conventional nonlinear dimensionality reduction methods \cite{wu_structure-aware_2018}. 
Besides, these manifolds might intersect with each other as often observed in astronomical data \cite{libeskind_tracing_2018}.

The presence of noise and outliers is known to affect the results of manifold learning and dimensionality reduction techniques \cite{wu_structure-aware_2018,hein_manifold_2007}. 
Several solutions for denoising manifolds, such as structure-aware data consolidation \cite{wu_structure-aware_2018} and manifold denoising \cite{hein_manifold_2007} do not remove any points but instead move the points towards the manifolds. 
These techniques improve the results of clustering or manifold learning as long as the noise points are close to the manifolds.
However, noise points or outliers in the background far away from the manifolds are not distinguishable by them. 

Graph-based methods combined with Markov chain used for clustering and dimensionality reduction 
are mostly based on the computation of several dominant eigenvectors 
\cite{roweis_nonlinear_2000,coifman_diffusion_2006,dixit_introducing_2019}. 
Therefore, on large datasets with many samples building the transition matrix and computing the eigenvectors is both memory and time-consuming. 
Even methods such as Arnoldi iterations \cite{saad_numerical_2011} requires a substantial amount of resources 
to estimate dominant eigenvectors. 
Furthermore, these methods consider the case of a single or multiple manifolds in a medium without noise \cite{coifman_diffusion_2006,valencia-aguirre_multiple_2011}, 
which is not true in many physical applications, such as in cosmological simulations \cite{libeskind_tracing_2018}.
Several papers have addressed the problem of manifold denoising and structure discovery 
specifically for astronomical datasets \cite{libeskind_tracing_2018,shivashankar_felix:_2016,sousbie_Disperse_2011-1,sousbie_Disperse_2011_2}. 
In \cite{sousbie_Disperse_2011-1,sousbie_Disperse_2011_2}, and \cite{shivashankar_felix:_2016} the one-dimensional manifolds are detected based on topological structure and Delaunay tessellation of the data. 
These techniques typically do not specify whether points belong to one-dimensional manifolds, but determine the medial axis.
Extensions to higher dimensions are computationally expensive.

Noise is also known to adversely affect other scenarios, such as clustering, and hence several strategies have emerged to deal with that problem in that context.
Clustering With Outlier Removal \cite{liu_clustering_2021} deals with this challenge by combining K-means clustering in partition space with Holoentropy. 
While being an interesting approach several disadvantages limit its use in practice, namely:
the user needs to provide the number of outliers in advance, and due to the use of K-means clustering, it is not suitable for elongated or curved clusters.
The Longest Leg Path Distance (LLPD) \cite{little_path-based_2020} was proposed as a new metric for elongated noisy clusters, which is used to remove points if the metric value is larger than a predetermined threshold.
This metric was demonstrated to be beneficial for clustering data with elongated clusters in spectral clustering \cite{little_path-based_2020}.
However, it provides no solution 
if clusters are highly curved and of different sizes. 
Hierarchical Clustering algorithm Based on Noise Removal (HCBNR) \cite{cheng_hierarchical_2019} reduces the number of parameters and 
noise by defining natural neighbors, which enable 
the discovery of arbitrary-shape clusters. 
And the Adaptive Density-Based Spatial Clustering (ADBSCAN) \cite{li_novel_2020} does not require prior knowledge about the number of clusters and 
introduces a directed nearest neighbourhood graph to denoise the data.
However, both ADBSCAN and HCBNR consider a point as noise if the density of their 
neighbourhood is smaller than a threshold. 
Therefore, 
they often fail to separate clusters with different densities and 
may lose parts if the density varies along the cluster.
Moreover, approaches for multiple manifold clustering show limitations dependent on the number of noise points present
\cite{deutsch_intersecting_2015,gong_robust_2012}. 
The latter suggests to handle this problem by removing outliers only if they constitute a small fraction of the number of inliers, but explicitly 
state that it is 
not extendable for many noise points. 

The problem of detecting points that belong to specific structures/manifolds, while simultaneously identifying noise points in large datasets, 
can be viewed as an optimisation problem.  
From this viewpoint, it is similar to a standard swarm based optimisation problem where the aim is to discover the shortest path between all routes.
The ant system \cite{dorigo_ant_1996} and ant colony system \cite{gambardella_solving_1996} are two heuristic methods applied to problems including the travelling salesman, vehicle routing and job scheduling. 
Several studies examined the ants' behavior in a colony for clustering data with a modest amount of outliers \cite{fahy_ant_2019, menendez_medoid-based_2016}. 
In \cite{fahy_ant_2019}, they proposed a computationally efficient method based on the behaviour of a certain type of ants to cluster streaming data that is also scalable  
to higher dimensions.
Both methods mostly perform density-based clustering and hence underperform in denoising and clustering lower-dimensional manifolds with significant curvature. 
Moreover, they studied a limited proportion of noise with respect to the data points, which is often not realistic in real-world scenarios.

In this contribution, we propose the Locally Aligned Ant Technique (LAAT)\footnote{Code and material available at 
\url{https://github.com/abst0603/LAAT}}, a novel extension to the ant colony algorithm able to remove noise from large point clouds and capture multiple potentially faint structures 
buried within. 
We demonstrate that depositing pheromone and an pheromone evaporation mechanisms is useful and effective in ignoring outliers and finding points on multiple structures. 
Our experiments confirm that the formulation of the ant's jumping preference, based on the alignment with the manifold tangent vector, 
outperforms the pure density based decision based on the Euclidean distance between a point and its neighbours.  
Besides, 
Furthermore, we demonstrate 
the proposed method on astronomical data with 6.7 million samples, showing that our method is capable of separating the data 
meaningfully. 
Note that this is a different problem compared to typical image processing and computer vision tasks \cite{Pless2009}.
There, the emphasis is on capturing a limited number of natural degrees freedom that characterise image collections or video streams. 
In contrast, we aim to zoom on potentially intricately related complex low-dimensional manifold structures embedded in spaces of much lower dimension than that of the image spaces.
Therefore the proposed method is neither a traditional clustering nor manifold learning method. 
It is rather a denoising strategy that improves the result of 
such techniques in multiple scenarios: 
1) 
significant presence of outliers, 
2) multiple elongated clusters or nonlinear manifolds, and 
3) 
structures with varying noise within and across. 
Demonstrating the LAAT as preprocessing for clustering and manifold learning is out of the scope of this paper, and corresponding experiments are provided as supplementary material in the code Git repository.

This paper introduces the extended ant algorithm and its theoretical foundation in section \ref{sec:method}.
In section \ref{sec:data} synthetic and astronomical datasets demonstrate the parameter robustness and performance of LAAT compared to other techniques.
A concrete example extending structure detection in higher dimensional data is given in section \ref{high-dimension-section}, an analysis of parameter in section \ref{paramSet},
and conclusions are found in section \ref{sec:conclusions}.

%
%
%
%

\section{Extracting manifolds using the ant colony}
\label{sec:method}
The ant colony algorithm is traditionally 
used to detect the shortest path between the original starting point and the destination, where there are many routes between these two points. 
Generally such ant colony algorithms implement a random walk procedure in which simulated agents 
move from point to point in search of a reward (the artificial counterpart to food) guided by two factors:
\begin{enumerate}
    \item the heuristic for choosing the shortest path, which is sometimes the inverse of the route length between the ants' nest and the food \cite{dorigo_optimization_1992}, 
    \item the amount of pheromone $F^{i-j}(t)$ accumulated on the path between $i$ and $j$ at time $t$ on an ants path \cite{dorigo_new_1999}.
\end{enumerate}
These two factors incentivize each agent to select a short path used by many ants reinforcing its use further. 
Inspired by nature, it is proposed in \cite{dorigo_new_1999} to gradually evaporate the pheromone on all routes: 
\begin{equation}
F^{i-j}(t+1) = (1-\zeta)\cdot F^{i-j}(t) \enspace ,
\label{evaporationformula}
\end{equation}
where $0<\zeta<1$ denotes a constant evaporation rate to be chosen in advance. 
As a consequence, routes which are chosen by fewer ants tend to fade faster over time until finally, one path emerges as the preferred path. 
In contrary to the algorithms stated above, our aim is not finding the shortest path, but separating points which belong to potentially low-dimensional manifold structures embedded in high-dimensional noisy points clouds. 
To achieve the goal, we introduce a novel ant colony strategy including a directional preference to move to points aligned with dominating directions of their local neighborhood enabling the discovery of manifolds,  
as detailed in the following. 

\subsection{Alignment based ant colony algorithm}
\label{sec:AntColonyAlgo}
We consider a data set $\mathcal{D} = \{ \vec{x}_1, \vec{x}_2, \dots, \vec{x}_n \}$ of $n$ points $\vec{x}_i \in \mathbb{R}^D$. 
For points in $D$-dimensional space, there are $D$ principle components in $\mathcal{N}^{(i)}_r$, a neighborhood with radius $r$ centered at point 
$\vec{x}_i$. 
Denote the local eigenvectors and eigenvalues by $\vec{v}_d$ and $\lambda_d$ for $d=1,2,...,D$, respectively. 
When jumping from $\vec{x}_i$ to one of its neighbors $\vec{x}_j \in \mathcal{N}^{(i)}_r$, we would like to give higher preference to jumps along the dominant eigen-directions ($\vec{v}_d$ with high $\lambda_d$).
Local PCA performed on $\mathcal{N}^{(i)}_r$ yields, under the assumption of distinct 
eigenvalues, unique unit eigenvectors (up to possible direction flips indicated by opposite signs). 
We quantify the alignment of the jump vector $(\vec{x}_j-\vec{x}_i)$ with $\vec{v}_d$ 
through their angle $\alpha_{d}^{(i,j)}$,
in particular $|\cos \alpha_{d}^{(i,j)}|$.
The alignment values are then normalized over all dimensions to give relative weighting of the alignment of $(\vec{x}_j-\vec{x}_i)$ with the eigen-directions by
\begin{equation}
    w_{d}^{(i,j)} = 
    \sfrac{|\cos \alpha_d^{(i,j)}|}{\sum\limits_{d^\prime=1}^D |\cos \alpha_{d^\prime}^{(i,j)}|} \enspace .
    \label{cosineweight}
\end{equation}
Likewise, we express the relative importance of different eigen-directions through normalized eigenvalues
\begin{align}
    \overline{\lambda}_{d}^{(i)} = 
    \sfrac{\lambda^{(i)}_d}{\sum\limits_{d^\prime=1}^D \lambda^{(i)}_{d^\prime}} \enspace .
    \label{normalizeeigenvalues}
\end{align}
We now define the preference for moving from $\vec{x}_i$ to $\vec{x}_j$ as 
\begin{equation}
E^{(i,j)} = \sum\limits_{d=1}^D w_{d}^{(i,j)} \cdot \overline{\lambda}_{d}^{(i)} \enspace, 
\label{Eformula}
\end{equation} 
meaning that it is more desirable to jump from $\vec{x}_i$ to $\vec{x}_j$ if $(\vec{x}_j-\vec{x}_i)$ is aligned with dominant eigen-directions. 
We normalize this preference within $\mathcal{N}^{(i)}_r$ (the neighborhood of $\vec{x}_i$) resulting in the relative 
preference: 
\begin{equation}
\overline{E}^{(i,j)} =
\sfrac{E^{(i,j)}}{\sum\limits_{j^\prime\in \mathcal{N}^{(i)}_r} E^{(i,j^\prime)} } \enspace .
\label{Enormalizeformula}
\end{equation}
Considering the example of uniform or Gaussian noise, which exhibits roughly equal eigenvalues in all directions, the equations \eqref{Eformula} and \eqref{Enormalizeformula} result in higher probabilities of jumping to manifold points since they will provide the dominant eigen-directions aligned with the structure. 
This process gradually motivates the concentration of pheromone on the points alongside the manifold, rather than the points in background noise.

We define a slightly modified notion of the biologically inspired pheromone with $F^j(t)$ being the amount of pheromone accumulated at point $\vec{x}_j$ at time $t$. 
The relative pheromone is normalized within the local neighborhood of $\vec{x}_i$:
\begin{equation}
\overline{F}^j(t) = \sfrac{F^j(t)}{\sum\limits_{j^\prime\in \mathcal{N}^{(i)}_r} F^{j^\prime}(t) } \enspace .
\label{pheromonenormalized}
\end{equation}
For social network graphs \cite{chen_ant_2020} proposed to deposit pheromone on nodes to mark the ones propagating unwanted information.
Similarly, we refrain from pheromone being released on the path between pairs of points in favor of its accumulation on the nodes themselves.
For our application, this has two advantages: 
it is straightforward to reason which points may belong to a manifold based on the amount of the pheromone accumulated on them, and it is computationally more efficient.

We now combine the alignment with the pheromone to define the  movement preference from $\vec{x}_i$ to $\vec{x}_j$ of an ant by:
\begin{equation}
V^{(i,j)}(t) = (1-\kappa)\overline{F}^j(t)+\kappa \overline{E}^{(i,j)} \enspace. 
\label{vformula}
\end{equation}
The parameter $\kappa \in [0,1]$ tunes the relative importance of the influence of the alignment and pheromone terms. 
In analogy with statistical physics, we view the jump preferences $V^{(i,j)}(t)$ as negative ``energies'' associated with the jumps and express the jump probabilities as:
\begin{equation} 
P(j|i,t) = \frac{\exp(\beta V^{(i,j)}(t))}{\sum\limits_{j^\prime\in \mathcal{N}^{(i)}_r} \exp(\beta V^{(i,j^\prime)}(t))} \enspace ,
\label{pformula}
\end{equation}
where $\beta>0$ plays the role of the inverse ``temperature''. 

We now summarize the proposed ant algorithm for the detection of potentially low-dimensional manifolds embedded in high-dimensional point clouds (pseudo code provided in Algorithm \ref{algo}
). 
Hyper-parameters such as the number of ants $N_\mathrm{ants}$, epochs $N_\mathrm{epoch}$ and steps $N_\mathrm{steps}$ are chosen according to the available memory and computational resources of the user. 
As generally true for MCs, the more the better.
Intrinsic parameters to be selected are the neighborhood radius $r$, inverse temperature $\beta$, tuning between alignment and pheromone preference $\kappa$ and deposited amount of pheromone $\varphi$, and evaporation rate $\zeta$. 
They allow data and application specific tuning to structures the user is interested to find. 
They are very robust in practice as shown in section \ref{sec:experiments}. 
Then, the local neighbors of $\vec{x}_i$ are detected, and we compute local PCA for each point and save the results to save computational cost (see section \ref{sec:costs}). 
To urge the ants to start a path from a denser neighborhood the random starting points should satisfy:
\begin{equation}
\label{startingcond}
    |\mathcal{N}^{(i)}_r| >= \widetilde{\mathcal{S}}
\end{equation}
where 
$\widetilde{\mathcal{S}}$ denotes the median of the set $\mathcal{S}=\{|\mathcal{N}_r^{(i)}|\ |  \vec{x_i} \in \mathcal{D}\}$. 
Every ant performs a random walk on points from ${\cal D}$ based on jump probabilities \eqref{pformula} for $N_\mathrm{steps}$. 
The indices of the visited points by ant $k$ are stored in the route multiset $R^{(k)}$. 
At the end of the walk the pheromone is updated according to:
\begin{equation}
\label{pheromoneupdate}
F^{j}(t) = F^{j}(t-1)+\nu (j) \varphi, \qquad \forall j\in R^{(k)},
\end{equation} 
where $\varphi$ denotes the constant amount of pheromone deposited on a point visited by an ant and $\nu(j)$ is the multiplicity of element j in multiset $R^{(k)}$.

A central part of the method is pheromone evaporation based on \eqref{evaporationformula} that is applied to all points (rather than the path connecting them) after all ants finish their routes attaching pheromone at each step on each visited notes. 
Equation \eqref{pformula} with $\kappa\neq 1$ exhibits a reinforcing dynamic 
that gradually increases the probability to visit manifold nodes and hence the amount of pheromone accumulated on them in comparison to noise points.
While pheromone evaporation attenuates the amount on all the points it enhances this reinforcement since it lessens the visitation probability for rarely visited points. 
\begin{algorithm}[t]
\caption{LAAT 
for discovering manifolds inside high-dimensional noise data points}
\label{algo}
{\bf Input:} Data $\mathcal{D}$, $N_\mathrm{epoch}$, $N_\mathrm{ants}$, $N_\mathrm{steps}$, $r$, $\varphi$, $\zeta$, $\beta$, $\kappa$ \\
\nl $F^{i}(0) = 1$ $\forall x_i\in \mathcal{D}$ and initialize the parameters\\
 \For{$i = 1$, \dots, $n$}{
\nl     Find the local neighborhood $\mathcal{N}^{(i)}_r$ for $\vec{x}_i$   \\
\nl     Compute local PCA for each point    \\
\nl     Remove $x_i$ if $|\mathcal{N}^{(i)}_r| < d$ \\
}
\nl Compute the $\widetilde{\mathcal{S}}$\\
\For{$c=1$ to $N_\mathrm{epoch}$}{
    \nl     Place ants on random points as defined in (\ref{startingcond})\\
    \For{$k=1$ to $N_\mathrm{ants}$}{
        \For{$s=1$ to $N_\mathrm{steps}$}{
            \nl From current node $i$ select next node $j$ with probability as defined in \eqref{pformula}
        }
        \nl Update pheromone as defined in (\ref{pheromoneupdate}) \\
    }
    \nl Apply evaporation as defined in \eqref{evaporationformula} \\
}
\nl \Return $\vec{F}$ containing amount of pheromone on nodes
\end{algorithm}

\subsection{Comparison with Markov Chain}
\label{CompMC}

In our proposed ant algorithm the agents release pheromone on their way jumping from node $i$ to $j$ with a probability determined by two weighted terms:
1) the alignment term, based on the eigen-vectors and values in the local neighborhood of $\vec{x}_i$, and 
2) the pheromone accumulated on the nodes from previous visits. The latter baring the consequence that the probabilities \eqref{pformula} change over time. 
However, if the ants ignore the pheromone ($\kappa=1$) their next step just depends on the direction of the principal components in the local neighborhood and thus the transition probabilities will stay constant over time. 
This special case can be considered as a Markov Chain (MC) with a fixed transition matrix and can be studied as a stationary distribution over the point cloud. 
Therefore, the probability of jumping from point (state) $x_i$ to another one $x_j$ is determined by $\overline{E}^{(i,j)}$ independent of the time $t$, 
and hence each state only depends on the previous state:
\begin{equation}
P(j|i) = 
\frac{\exp\left(\beta \overline{E}^{(i,j)}\right)}{\sum_{j^\prime\in \mathcal{N}^{(i)}_r} \exp\left(\beta \overline{E}^{(i,j^\prime)}\right)
} \enspace.
\label{markovpformula}
\end{equation}
Therefore, the probability of jumping from point $\vec{x}_i$ to $\vec{x}_j$ is higher than 
jumping to $\vec{x}_k$ if the difference vector $\vec{x}_j - \vec{x}_i$ is more aligned with the main eigenvector than  $\vec{x}_k - \vec{x}_i$ is.

If the neighborhood radius is large enough the set of points in our dataset will form a  recurrent class, so that a path exists between any pair of points in the neighborhood graph. Under some additional assumptions, the point cloud can be identified with an ergodic Markov Chain (EMC) \cite{grinstead_introduction_1997} characterised by the transition matrix $P_{ij}=P(j|i)$. 
In case the point cloud is organised in several isolated point clusters, we will have a set of distinct EMCs that can be analysed in isolation.

One key property of an EMC is that 
it converges to a steady-state which does not depend on the starting point, but only depends on the constant probabilities of jumping between the points \cite{grinstead_introduction_1997}.
Therefore, an EMC will converge to the steady state vector, which contains asymptotic visitation frequencies  of all states (points). This steady-state vector $\pi$ is the dominant left eigenvector of the transition matrix \cite{grinstead_introduction_1997} and we have for all initial states $i$:
\begin{equation}
\lim_{s\to\infty} P_{ij}^s = \pi_j,
\label{pijn} 
\end{equation}
{where $P^s$ is the $s$-th power of $P$ and $P_{ij}^s$ is the probability of jumping from point $i$ to point $j$ in $s$ steps.} 
The transition matrix $P$ grows quadratically in size with the number of points and therefore the computation of the eigenvectors quickly becomes computationally prohibitive to calculate for large datasets. 
Therefore, we take advantage of the Power method for computing the left dominant eigenvector \cite{powermethod}. 
This vector contains the visitation frequency for each point and assuming manifold points are more frequently visited we infer 
which points are more likely to belong to a manifold 
by thresholding. 

The idea of discovering the underlying manifold inside a point cloud using a MC has previously been presented for dimensionality reduction and clustering applications \cite{wu_structure-aware_2018,little_path-based_2020,dixit_introducing_2019,steinerberger_filtering_2016}. 
Those papers define the transition matrix according to a distance measure on the data capturing dense regions independent on dominant directions and alignments of manifolds. 
However, the stationary distributions of these MCs often do not point sufficiently strongly to the desired set of manifold-aligned points.
In contrast 
our new definition captures directional information of the data, which can reveal manifold structures embedded in large amounts of noise points in high dimensions. 
Furthermore, the incorporation of pheromone accumulation and evaporation, inspired by natural ant systems, reinforces areas visited frequently and diminishes noise were the pheromone degrades through time.
Our methodology effectively modifies the MC structure (transition probabilities) in every iteration, and hence a theoretical analysis is extremely challenging and well beyond the scope of this paper.
We instead focus on detailed empirical evaluations and analysis of the hyper-parameters.
%
%

\subsection{Computational complexity of LAAT}\label{sec:costs}
The complexity of the LAAT algorithm depends on the number of data points $n$, their dimensionality $D$, and the choice of the hyperparameters. 
The latter involves a data dependent component, namely the number of points in the neighborhood $|\mathcal{N}_r^{(i)}|$ with radius $r$ of each sample $\vec{x}_i$. 
We will state the worst-case upper bound with $\Delta=\max_i|\mathcal{N}_r^{(i)}|$ being the maximum number of neighbors for radius $r$. 
Equations \eqref{cosineweight}-\eqref{Enormalizeformula} 
can be computed as a preprocessing step to reduce computational complexity at the cost of memory as implemented in Algorithm \ref{algo}. 
The neighborhoods are naively computed in $O(n^2D)$, which can be sped-up using approximate techniques, such as \cite{blanco2014nanoflann}. 
For each neighborhood of maximum $\Delta$ points, local PCA is computed using singular value decomposition, which in general takes 
$O(\Delta^2 D + D^3)$ flops. 
This can of course change dependent on the particular implementation, we use
\cite{eigenweb}. 
Therefore the preprocessing amounts to $O(n^2 D + n(\Delta^2 D+D^3+\Delta D^2))$. 
Finally, the costs of the iterative procedure is maximally $O(\mathrm{N_{epoch}}\mathrm{N_{ants}}\mathrm{N_{steps}}\Delta)$. 
Note that the design using $\mathrm{N_{epoch}}$ and $\mathrm{N_{ants}}$ allows for parallelization.

\section{Data}
\label{sec:data}
We demonstrate our proposed method discovering potentially numerous manifolds of varying density in noisy backgrounds with several experiments on synthetic and real-world astronomical datasets.
Notably, we highlight the practical applicability of this method on a large scale particle simulation of the cosmic-web
\cite{libeskind_tracing_2018}, in which the galaxies and dark matter are distributed non-uniformly on many 1 and 2-dimensional manifolds (filaments and walls, respectively).

\subsection{Synthetic data}
Two synthetic datasets are designed with similar challenges as real datasets, namely highly curved manifolds, with 
varying density on the manifolds itself (two-arms dataset), and multiple manifolds of 
different density. 
They serve as ground truth to compare 
LAAT with existing methods and parameter analysis. 

\begin{figure}[t]
    \includegraphics[width=\columnwidth]{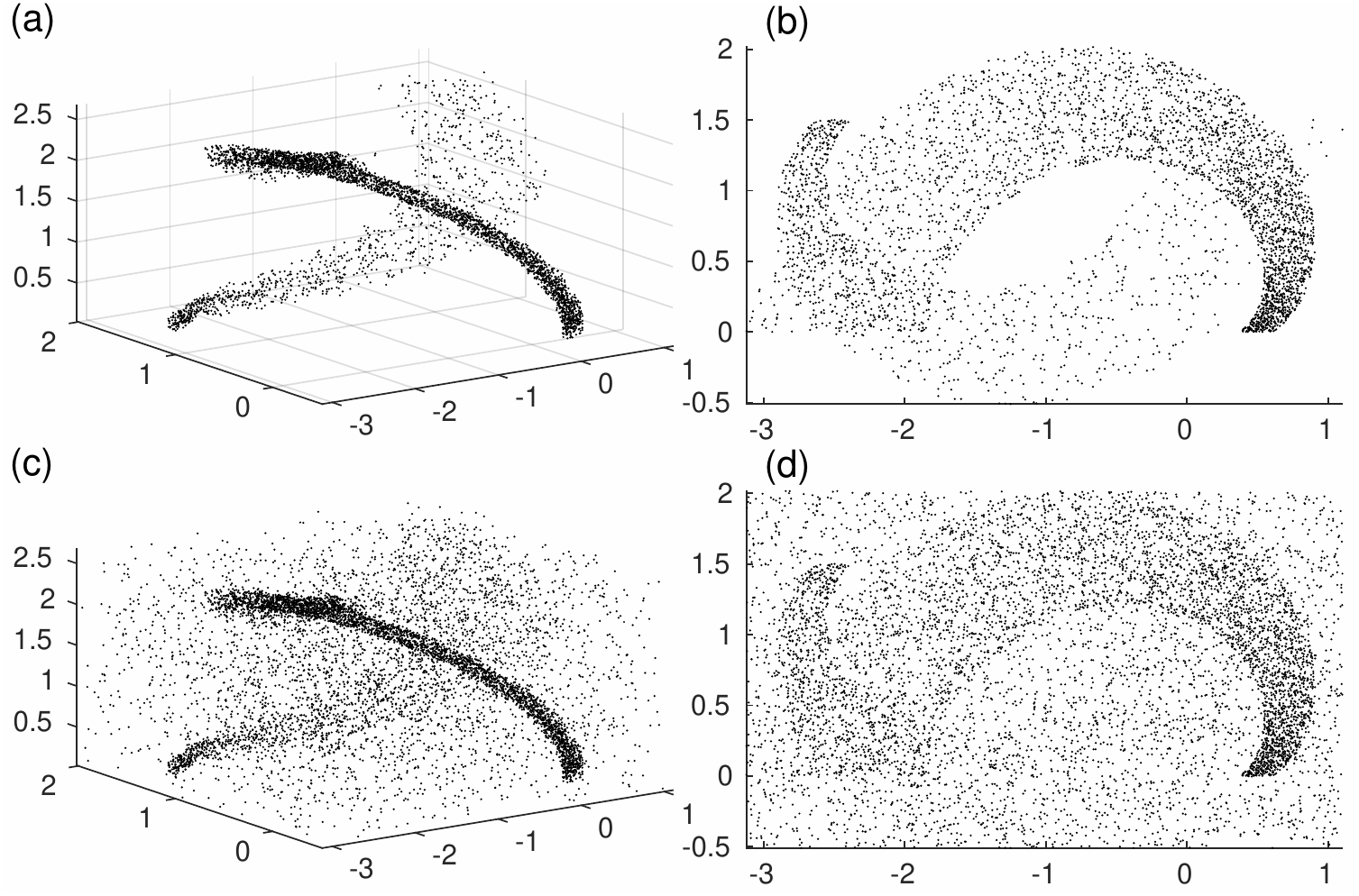}
    \caption{3D (a,c) 
    and front views (b,d) of the two-arm dataset, shown without (top) and with background noise (bottom).}
    \label{fig1}
\end{figure}

\paragraph{Two-arms} 
This dataset contains two non-intersecting two-dimensional noisy arms curving in three dimensions positioned similar to a double helix embedded in 8000 three-dimensional uniform noise points (see Fig.\ \ref{fig1}). 
The design is loosely based on the form of astronomical objects such as tidal tails in galaxies, or ram pressure stripped gas wakes in cluster disk galaxies.
Therefore, the distribution of points on the two arms, with 3000 on one and 1000 points on the other, is most dense at the narrow end, denoting the most recent position of the object, and decreases along the path where the particles have spread along a plane curved and tilted in space. 
To add to the complexity, the two manifolds contain noise with deviation from the 2D surface, such that the thickness of the arms is not zero. 
Specifically, uniform noise $U[0,0.2]$ in orthogonal direction to the manifolds is added. 

\begin{figure}[t]
\includegraphics[width=\columnwidth]{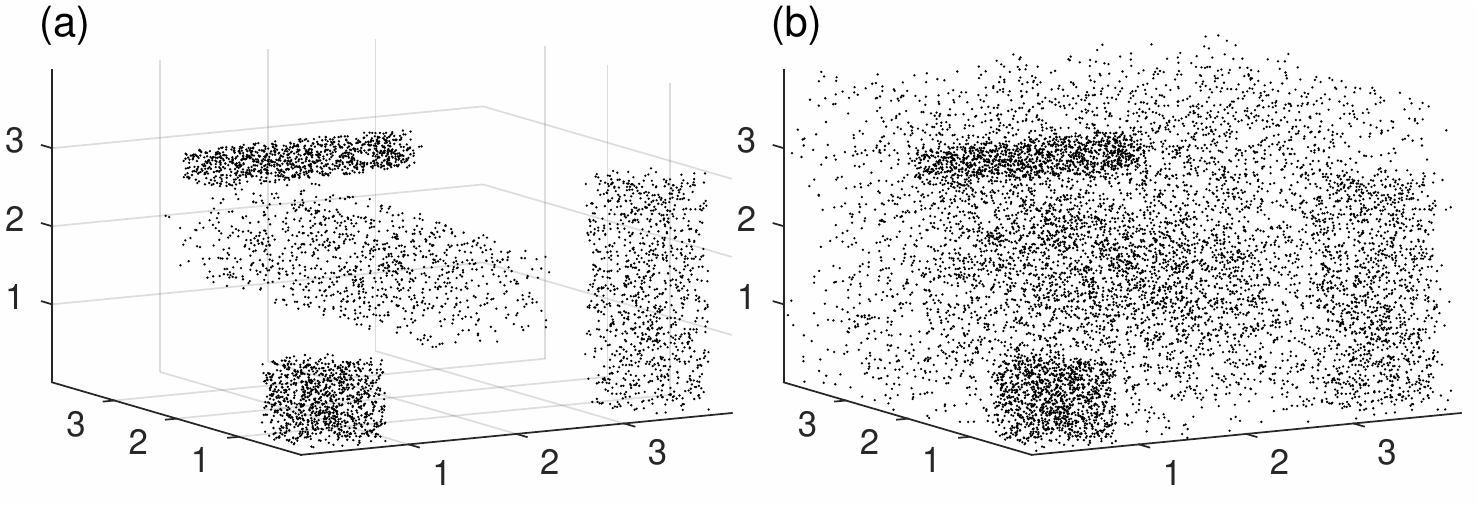}
\caption{Four 3D cylinders of varying density shown without and with uniform background noise (panel a and b).}
\label{fig2}
\end{figure}

\paragraph{Cylinders} 
This dataset contains four non-intersecting cylinders comprised of 1000 points each with varying height and diameter, resulting in a vastly different density for each of them. 
More precisely, the radii of two vertical cylinders 
are equal but their heights differ. 
5000 data points with a uniformly random distribution are inserted as background noise to the dataset. 
No orthogonal ``thickness'' noise was added to the manifolds.
Fig.\ \ref{fig2} visualizes the dataset before and after adding the background noise in panel (a) and (b). 

\subsection{Data from a Cosmological N-body simulation}
\label{subsection: cosmoligicalNBody}
On mega-parsec scales, galaxies, galaxy clusters, and dark matter are not uniformly distributed in space. 
Due to the presence of small perturbations in the density-field of the early Universe, gravity later accumulates matter into a variety of large scale structures, including walls (which are flat, plane-like structures), filaments (which are long thin structures that arise where walls cross each other), and clusters (which are more spherical clumps of matter, arising where filaments cross, and that form some of the densest cosmological environments).
Astronomers also observe gigantic and roughly spherical under-dense regions, called voids, containing few galaxies, whose boundaries are defined by the walls \cite{libeskind_tracing_2018,platen_cosmic_2007}. 
In order to study how galaxy evolution is connected to their location in this large scale structure, it is necessary to identify the filaments, walls and clusters either observationally, or in large volume N-body cosmological simulations. 
 
\begin{figure}[t] 
    \centering
    \includegraphics[width=\columnwidth,height=4.8cm]{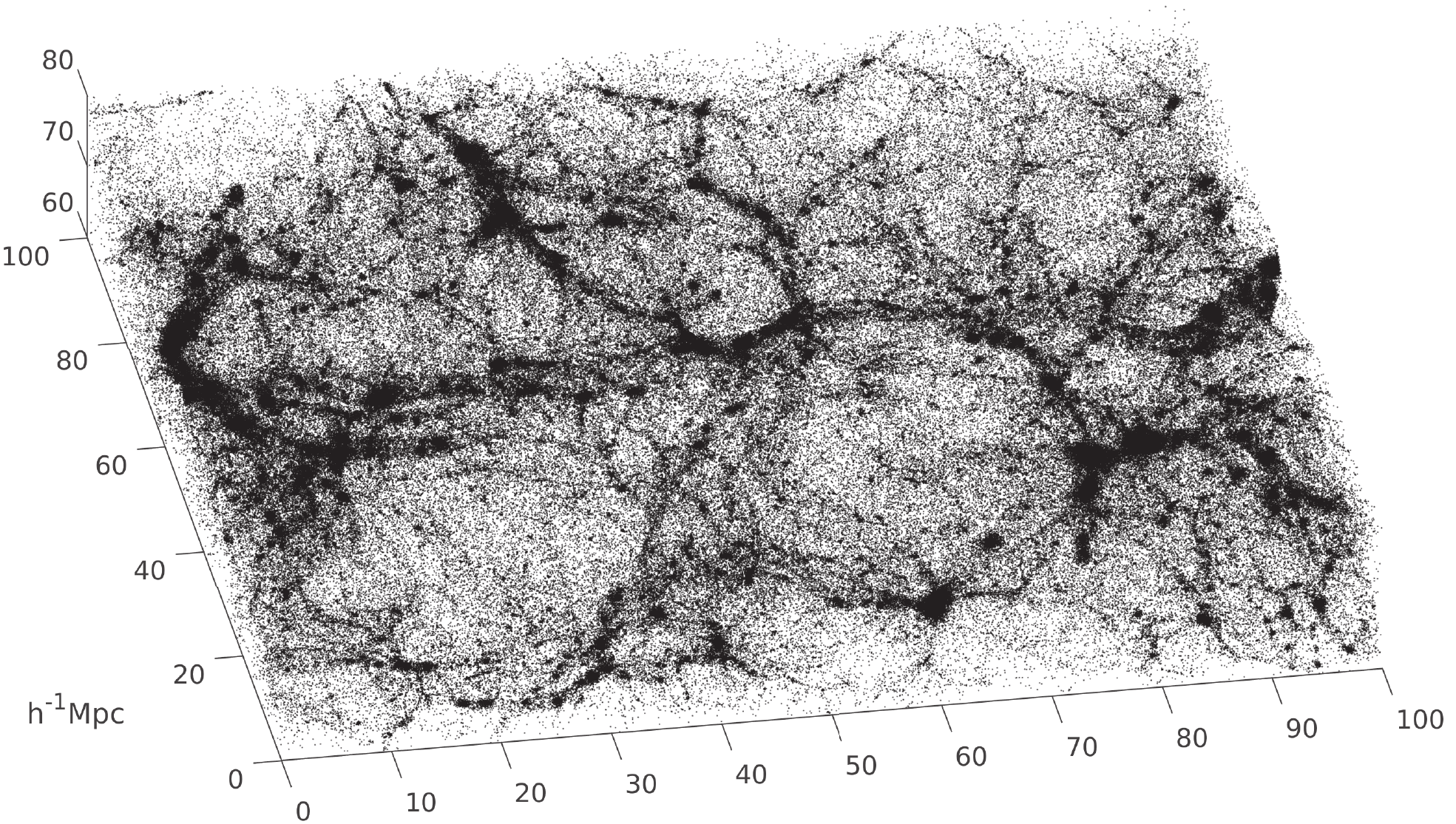}
    \caption{
    810K particles from a cosmic-web simulation.
    } 
    \label{fig3}
\end{figure}

Figure \ref{fig3} shows an example datacube of such a simulation. 
Our example simulation consists of 6.7 millions data points in a cube with edge size of 120$h^{-1}\mathrm{Mpc}$.
Similar to observed galaxies the points accumulate on a web-like structure composed of multiple filaments, walls and void structures, with varying sizes and densities. 
In such an N-body simulation, the complexity of the structure means there is no single, trivial way to define which particles belong to which large scale structures. 
As a result, there is no ``ground-truth'' 100$\%$ correct answer to the question which particle belongs to which structure.
This is a disadvantage for comparing our algorithm to other methods, which we mitigate 
by an additional experiment. 
 
To construct controlled experiments of this kind, we use three synthetic datasets with low-dimensional substructures that look similar to those in the full cosmological simulation. 
These were first introduced in \cite{shivashankar_felix:_2016} and are based on Voronoi Tessellation regions that mimic cosmological voids. 
Walls build the faces of the tessellation regions, which form filaments on the edges where they meet. 
Finally clusters are found at the corners where filaments meet, such that similar structures appear as can be observed in cosmological simulations. 
In contrast to the n-body simulations these synthetic data sets provide a ground-truth, allowing to identify exactly which data points are associated with a particular structure.
Note that the structures only appear morphologically similar to those in cosmological simulations and that they are unlikely to reproduce their more quantitative properties. 
Nevertheless, this enables us to quantify the relative performance of various methods for identifying substructures that, at least, broadly resemble those found in cosmological simulations. 
 
\begin{figure}[t]
    \centering
    \includegraphics[width=\columnwidth]{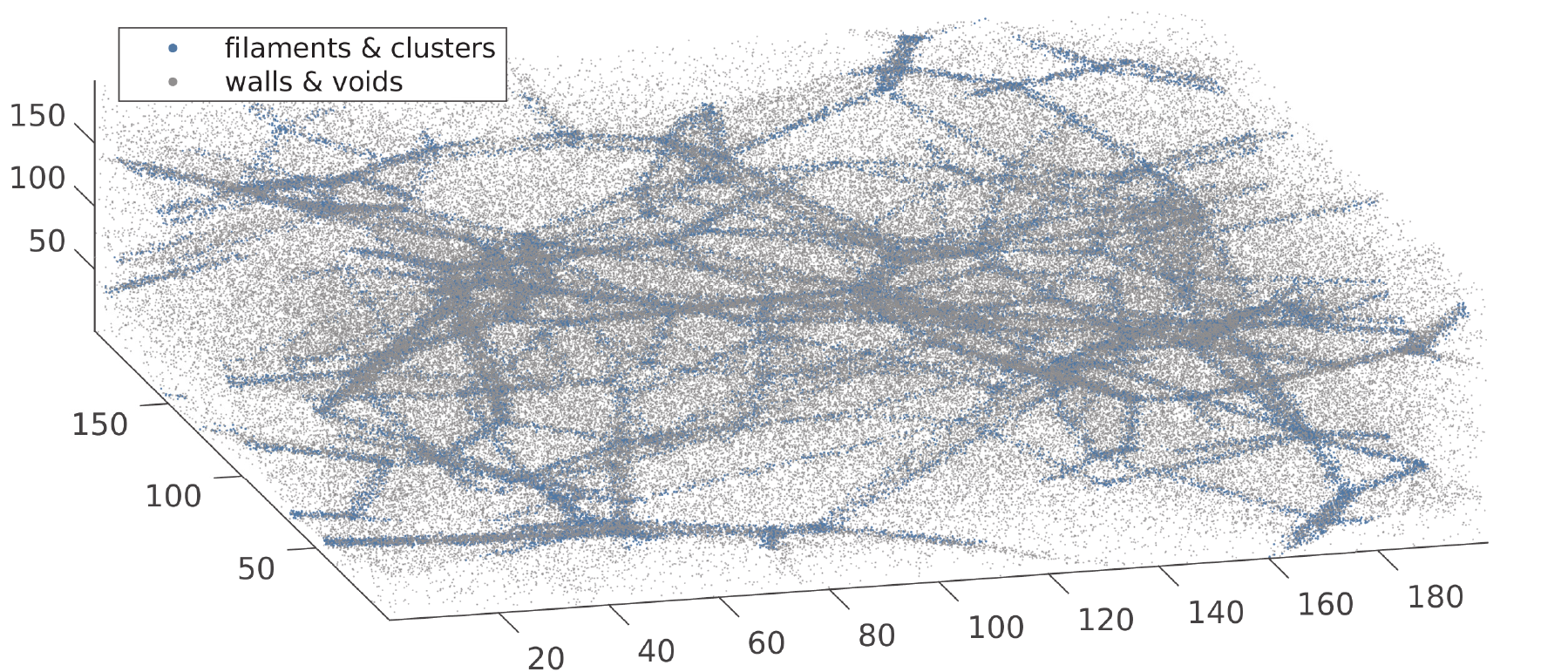}
    \caption{Synthetic cosmic-web 
    acquired by Voronoi tessellation.}
    \label{fig4}
\end{figure}

 Three synthetic datasets are generated each consisting of 262,144 particles formed by Voronoi tessellation around 32 centers inside a cube with an edge length of 200$h^{-1}\mathrm{Mpc}$. 
 The ratio of the number of points on filaments and clusters to walls and voids for datasets `A', `B', and `C' is 0.367, 1.324, and 6.254, respectively. 
 The cube of dataset `A' is presented in Fig.\ \ref{fig4}. 
 In contrast to N-body 
 simulations points in this idealised model for cosmological structures distribute uniformly on each filament, yet 
 their distribution varies across filaments. 

\section{Experiments and discussion}
\label{sec:experiments}
In this section, we first present a quantitative evaluation measure, namely the Hausdorff distance, for an objective comparison of 
techniques recovering manifolds from a noisy medium in the presence of a ground truth. 
Subsequently, we perform an empirical analysis of the parameters of the proposed LAAT 
based on the synthetic datasets as introduced before, and we investigate the benefit of pheromone in the procedure for 
detecting the manifolds.
Finally, we compare LAAT with Disperse, which is a well-known and often used method for detecting filaments in a synthetic and large n-body simulation of the cosmic web.

\subsection{Average Hausdorff distance}
The Hausdorff distance (HD) measures how far away from each other two non-empty sets $X$ and $Y$ in $\mathbb{R}^D$ are:
\begin{equation}
d_H(X,Y)=\max\!\left(\sup_{y\in Y}(\inf_{x\in X}
\lVert x-y \rVert),\sup_{x\in X}(\inf_{y\in Y}
\lVert x-y \rVert)\right)
\label{HausdorffDist}
\end{equation}
Here, we use the Euclidean distance between two points $x$ and $y$, which of course more generally could be any other distance \cite{boissonnat_geometric_2018}. 
The HD measures the largest of all distances between points on one set to their nearest neighbor in the second set. 
Nonetheless, because the HD computes the supremum of the distance among points, it is sensitive to outliers. 
To overcome this disadvantage, the authors in \cite{dubuisson_modified_1994} have suggested calculating the average distance instead of the maximum of the supremum distance. 
For two sets $X$ and $Y$ that contain $N_X$ and $N_Y$ number of points respectively, the average Hausdorff distance (AHD) is defined as: 
\begin{align}
d_\mathrm{AH}(X,Y)=
\dfrac{\displaystyle\sum_{y\in Y}\inf_{x\in X}\lVert x-y\rVert}{2N_Y}+
\dfrac{\displaystyle\sum_{x\in X}\inf_{y\in Y}\lVert x-y\rVert}{2N_X}
\label{HausdorffAvg}
\end{align}
consisting of two distinguished averaging parts with each part computing a one-side average distance between the two sets. 
The smaller the values for the Hausdorff distance become the closer the two sets are. 
Therefore, we use the AHD for datasets where we have a ground truth to compute how close the set of points $Y$ recovered from a noisy background by a manifold learning technique is to the true set of manifold points $X$.

\subsection{Numerical Experiments on Synthetic datasets}
\subsubsection{Influence of LAAT parameters}
\label{paramset}
In this experiment, we investigate the influence of the parameters on the final results of the LAAT algorithm. 
Parameters have mutual effect, such as $\kappa$, which determines the influence of the alignment and pheromone term, meaning the trade-off between following major directions versus reinforcing previous occupation, and $\beta$ which tunes the transition probabilities. 
Similarly, the effect of the amount of pheromone released by an ant is coupled to the evaporation rate. 
Therefore we empirically demonstrate the influence of these important parameters of 
LAAT 
performing a systematic parameter grid analysis using the two-arms dataset consisting of manifolds of varying density introduced before. 

\begin{figure}[t]
  \includegraphics[width=\columnwidth]{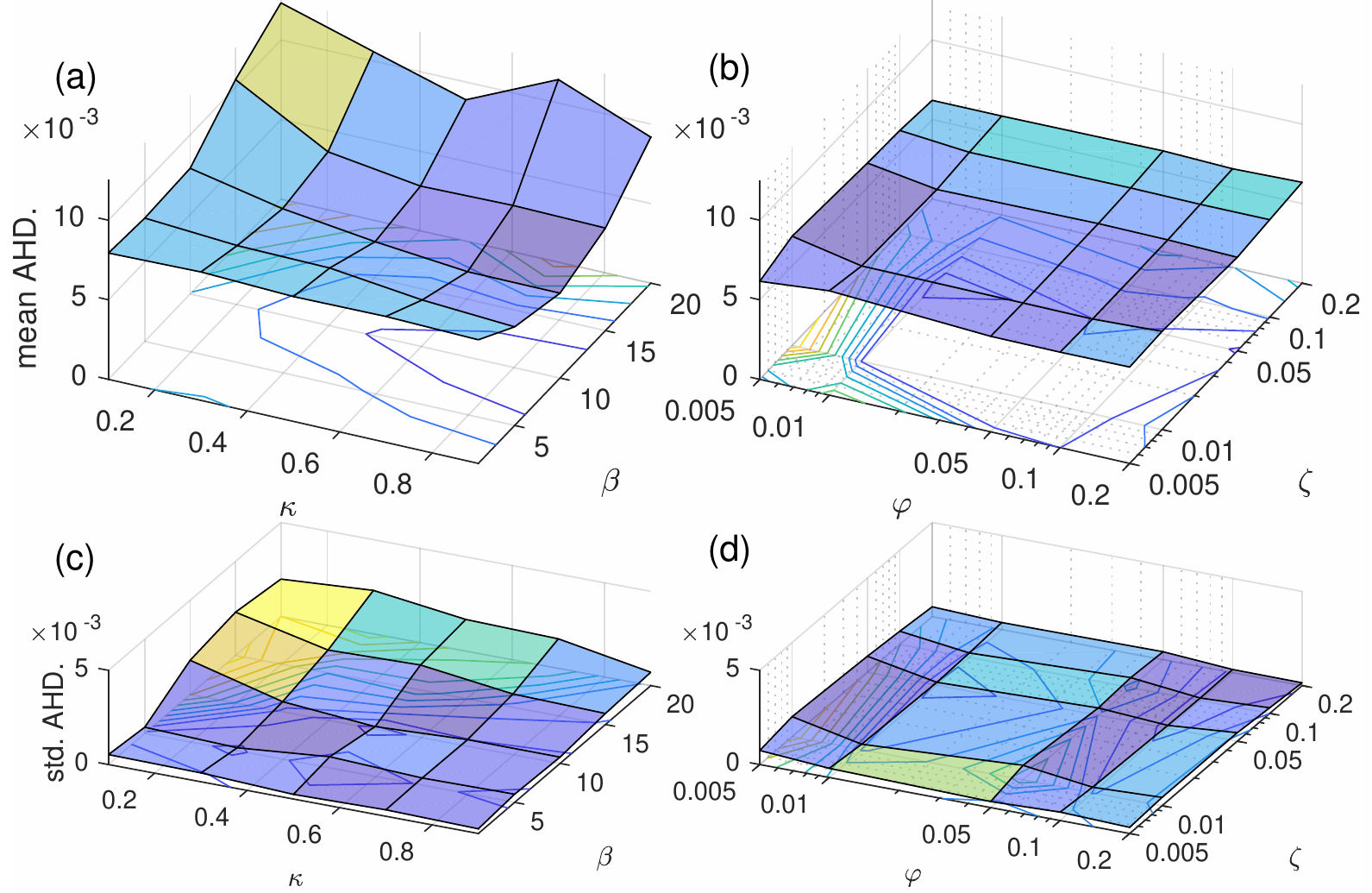}
  \caption{Two-arms: mean AHD and its standard deviation for parameter settings $\beta$ and $\kappa$ (panel a, c) and pheromone value versus evaporation rate (panel b, d) over 10 LAAT repetitions.
  }
  \label{paramfig}
\end{figure}

The amount of pheromone accumulated on the data samples is used to decide which points belong to a manifold or can be considered background noise. 
In real-world applications, including discovering filaments in the cosmic-web, one should therefore select a single value for the whole data as a threshold on the amount of pheromone on the points guided by domain knowledge. 
However, in this experiment, to make a fair examination of the results 
we derive a strategy to select the threshold and follow the same strategy for all ten repetitions of LAAT. 
To do so, we build a calibration set that contains the same two manifolds as the two-arm dataset with different background noise, namely 8000 new random points in the medium around the two manifolds as noise. 
For each repetition of the experiment with the same parameter setting we perform the following two steps: 
1) We first apply the LAAT strategy on the calibration set where the threshold on the resulting pheromone values is gradually reduced from maximum to the minimum across all points computing the AHD for each, respectively. 
2) The calibrated threshold value depicting the minimal AHD for the current parameter setup is then used as fixed threshold for all ten iterations on the original two-arm dataset. 
For most experiments parameters are initialized by 
default values, namely $\beta=10$, $\kappa=0.5$, $\varphi=0.05$, $\zeta=0.1$, $N_\mathrm{epoch}=100$, $N_\mathrm{steps}=2500$, and $N_\mathrm{ants}=100$.
For this dataset, we only reduced the number of epochs to $N_\mathrm{epoch}=20$.
Note that for the experiments of figure \ref{paramfig} panel (a) and (c), only the values of $\kappa$ and $\beta$ are changing, and in figure \ref{paramfig} (b) and (d), we alter the values of pheromone and the evaporation rate.
The resulting mean AHD and its standard deviation for the parameter couplings $\kappa$ and $\beta$ are shown in the left column of figure \ref{paramfig} (panel (a, c)), while corresponding values for the pheromone and evaporation rate are depicted in panel (b, d).

Figure \ref{paramfig} illustrates that the LAAT strategy recovers the underlying manifolds very well indicated by low AHD to the ground truth. 
It is furthermore robust against changing the $\beta$ and $\kappa$ (see panel (a) and (c)) in the range which the experiments were performed. 
There are points on the surface (a) corresponding to $\beta=10$ and $\kappa=0.9$, which depicts the lowest AHD with $5.3\times 10^{-3}$ on this dataset, however the difference to other settings is small. 
For $\beta$ equal to or larger than 20, the ants take a greedy policy following nearly exclusively the highest preference for the next jump, 
which generally results in larger standard variation across different repetitions and inclusion of more noise points away from manifolds.  
On the other hand, for $\beta$ smaller than 1 the ants decide randomly, thus more noise points survive after applying the threshold. 
By increasing the value of $\kappa$ from 0.1 to 0.9 for almost all values of $\beta$ the AHD decreases, which
indicates that the alignment is more important than the pheromone for detecting the manifolds with high curvature and varying noise. 

Figure \ref{paramfig} (b) and (d) demonstrate the effect of changing the pheromone amount $\varphi$ and evaporation rate $\zeta$ in the range [0.005,0.2]. 
Again the algorithm is very robust to changes in these parameters, since over the range which the experiment was performed it converges in all ten iterations to manifolds with very similar AHD. 
We observe that 
the amount of the deposited pheromone on each point increases by almost the same ratio for all points if we increase the $\varphi$ without much influence on the 
results.
Evaporation rate $\zeta>0.2$ 
results in 
more random jump decisions, as the pheromone on all points evaporates more quickly increasing the 
standard deviation.

\subsubsection{The discussion of a special case}
\label{phBenefit}
We demonstrate the benefits of the 
pheromone formulation using the four-cylinder dataset with multiple manifolds of varying density 
as introduced before. 
The pheromone serves as positive feedback reinforcing the chance of points on the manifold to get selected, 
since every time one ant visits a point it deposits extra pheromone on it and therefore motivates more ants to choose the same point.
As stated in section \ref{CompMC} for the special case of the ants not depositing any pheromone or ignoring it when deciding about their next step each jump is dependent on just the former one and the process can be analyzed as MC. 
In other words, the special case of $\kappa=1$ can be seen as MC including alignment information. 
Therefore, it is straightforward to prove that an EMC with a single class will converge to a steady state. 
However, as long as there are multiple classes/manifolds in a dataset, as is the case in 
the four-cylinder dataset, MC cannot always reveal all of them. 

Informally, the neighborhood radius influences the number of neighboring points around one ant that can be the next step of the ant. Thus, if the neighborhood is large, then there is a small chance for every point to get selected. 
As a result, the ants' decision will be more irregular, which typically 
leads to a more noisy output.
In this experiment, both the ant colony algorithm and the MC with the transition matrix based on alignment are compared for three different values of the radius of the neighborhood (namely 0.3, 0.4 and 0.5) on the four-cylinders dataset. 
Again we fix the parameters for the LAAT strategy to default values and correspondingly $\beta=10$ for the MC.
For the LAAT strategy we examine the behavior of ten independent runs while the MC result, being the vector of the steady-state of the system, can be computed analytically. 

\begin{figure*}[t]
  \includegraphics[width=\textwidth]{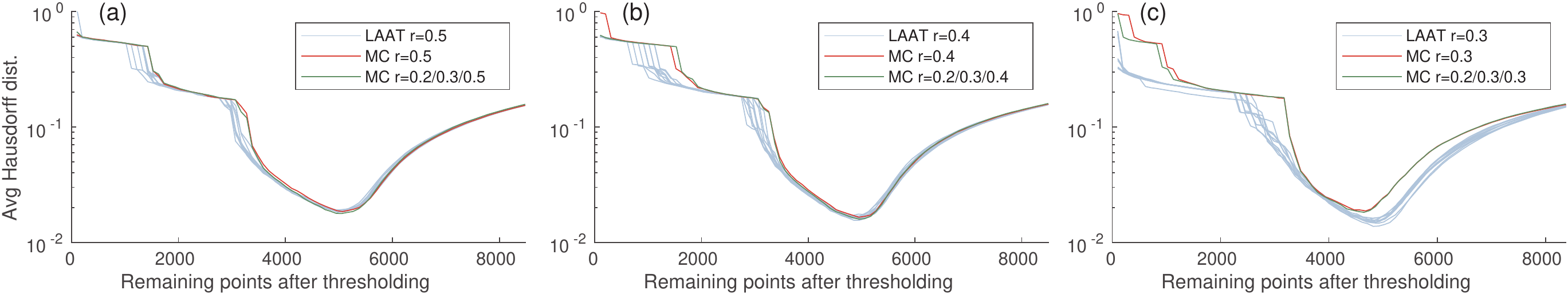}
  \caption{AHD curves for the ant colony algorithm, MC, and MC with altering radii. 
  Panels (a-c) show LAAT and MC performance with fixed neighborhood radius equal to 0.5, 0.4 and 0.3, respectively, and MC with altering radii as mentioned in the legend.
  }
  \label{AHDMCants}
\end{figure*}

Figure \ref{AHDMCants} illustrates the AHD on a logarithmic scale versus an increasing number of points remaining for all possible thresholds on the two deciding quantities, 
namely the amount of pheromone in LAAT and the steady-state vector corresponding to the visitation frequency in MC. 
In panel (a-c) reducing the threshold causes the number of points which survive to increases gradually, and the AHD continuously decreases until it reaches its minimum value, then it begins to increase monotonically until the threshold is smaller than the value of pheromone on all the points. 
The minimum AHD value for all LAAT runs and MC is found in the interval [4k 6k] exceeding the 4000 real manifold points, indicating that background noise, which is not on the manifolds is detected as well. 
As it is shown in 
figure \ref{AHDMCants}, the AHD between the ground truth and recovered manifolds is lower for the ant colony algorithm than MC on these three experiments for a range of thresholds, 
which reveals that LAAT generally recovers the manifolds better including fewer noise points. 

There are some sharp declines in the value of AHD in all figures, which corresponds to the discovery of the first point found on a new manifold. 
Figure \ref{AHDMCants} illustrates that the sharp declines happen faster in all LAAT runs than MC with the same radius of neighborhood. Since there are four cylinders with different sizes and densities inside the dataset, the MC with a transition matrix calculated based on a fixed radius (red lines in figure \ref{AHDMCants}) cannot converge on all of them.
To examine whether this is correct, we used the knowledge from ground truth about the position of points on or near to a cylinder. 
The new transition matrix was computed using different distances for the neighborhood of different points according to the knowledge we have from ground truth. 
If a point is near to or on the cylinder with the smallest radius, the neighborhood radius was set to 0.2. 
For the two cylinders with the same radius, the neighborhood radius was set to 0.3, and for the rest of the points including the cylinder with the largest radius, the value was set to 0.3, 0.4, and 0.5.
We therefore inform the MC which radius to use in which part of the dataset. 
As presented in the figure, the results of the informed MC 
are more similar to the results of ants, which shows that the ant colony algorithm with pheromone effect resembles the MC with altering radii. 
However, since the information about the position and number of manifolds buried in a noisy environment is typically not known in advance 
the ant colony algorithm, despite being computationally more expensive, is a very practical choice for unsupervised robust discovery.

\subsubsection{Convergence analysis of LAAT}
\begin{figure}[t]
    \centering
    \includegraphics[width=\columnwidth]{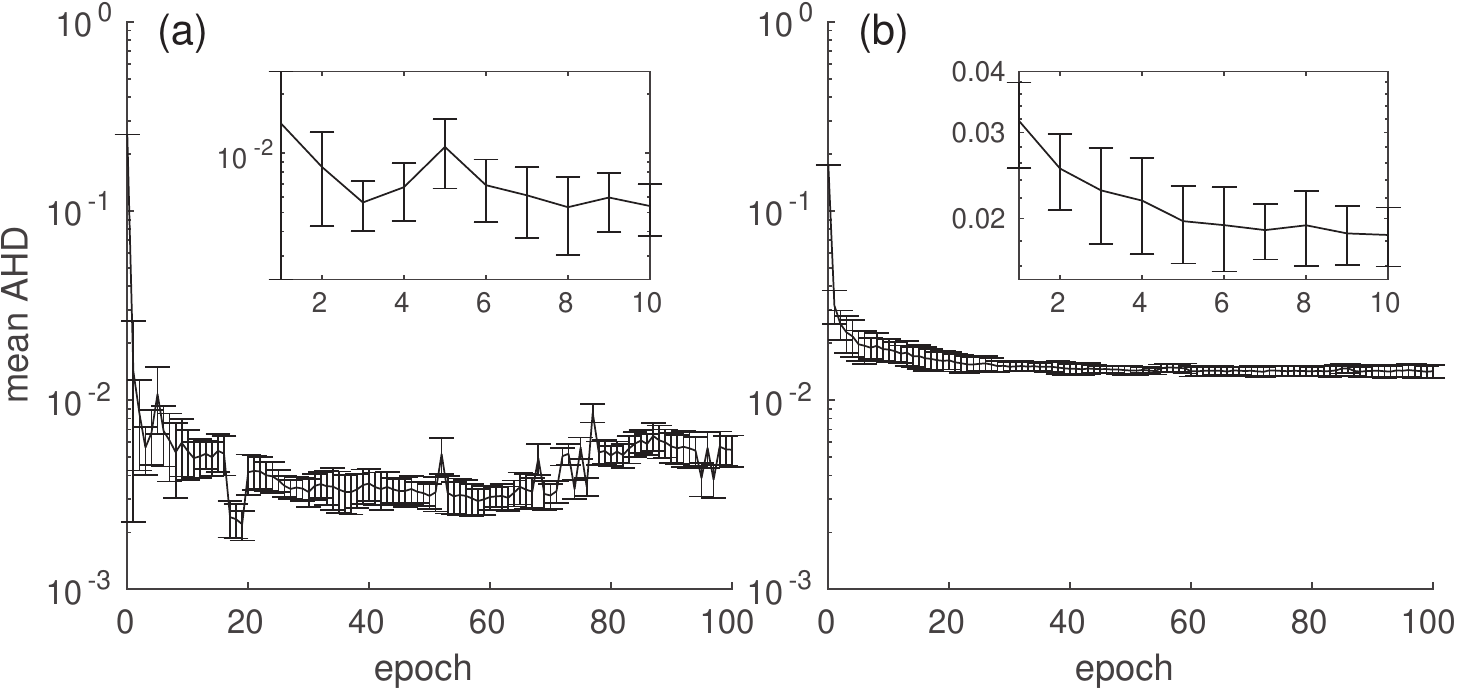}
    \caption{LAAT mean AHD curve of the (a) two-arm and (b) four-cylinder data and zoomed boxes showing epoch 1 to 10.}
    \label{convergenceplot}
\end{figure}

In this section, we investigate the convergence of the LAAT on the two-arm and four-cylinder datasets.
All parameters are initialized by the default values. 
The whole experiment is then iterated ten times. At each epoch, the threshold is selected automatically on a second dataset which contains similar manifolds but different random noise similar to the strategy which is described in the previous section. 
Figure \ref{convergenceplot} shows that the LAAT quickly converge to the final mean AHD in just 10 epochs (see zoomed region) 
on both datasets. As is stated in algorithm \ref{algo}, an epoch is started by sending ants one after another to explore the data points, and it is finished by pheromone evaporation.
We furthermore notice the small standard deviation across the independent runs indicating that although ants start from random points every epoch, all ants visit the manifold points more often and hence accumulate more pheromone there. 

\subsubsection{Synthetic dataset that broadly mimics cosmological structures (Synthetic cosmic-web)}
One proposed technique for identifying filaments inside the cosmic-web used by astronomers is the publicly available Disperse \cite{sousbie_Disperse_2011-1,sousbie_Disperse_2011_2}. 
Disperse 
builds a Delaunay tessellation field estimator density for each point. 
Then, using discrete Morse theory and theory of persistence it finds the clusters, filaments, and the walls and measures their different strengths. 
We compare the performance of the proposed LAAT method with Disperse on the three synthetic cosmic-web datasets \cite{shivashankar_felix:_2016} (see \ref{subsection: cosmoligicalNBody}). 
The difference between these datasets is the ratio of the number of particles on filaments and clusters (labeled as positive) as well as walls and voids (labeled as negative).

\begin{figure}[ht]
    \centering
    \includegraphics[width=\columnwidth]{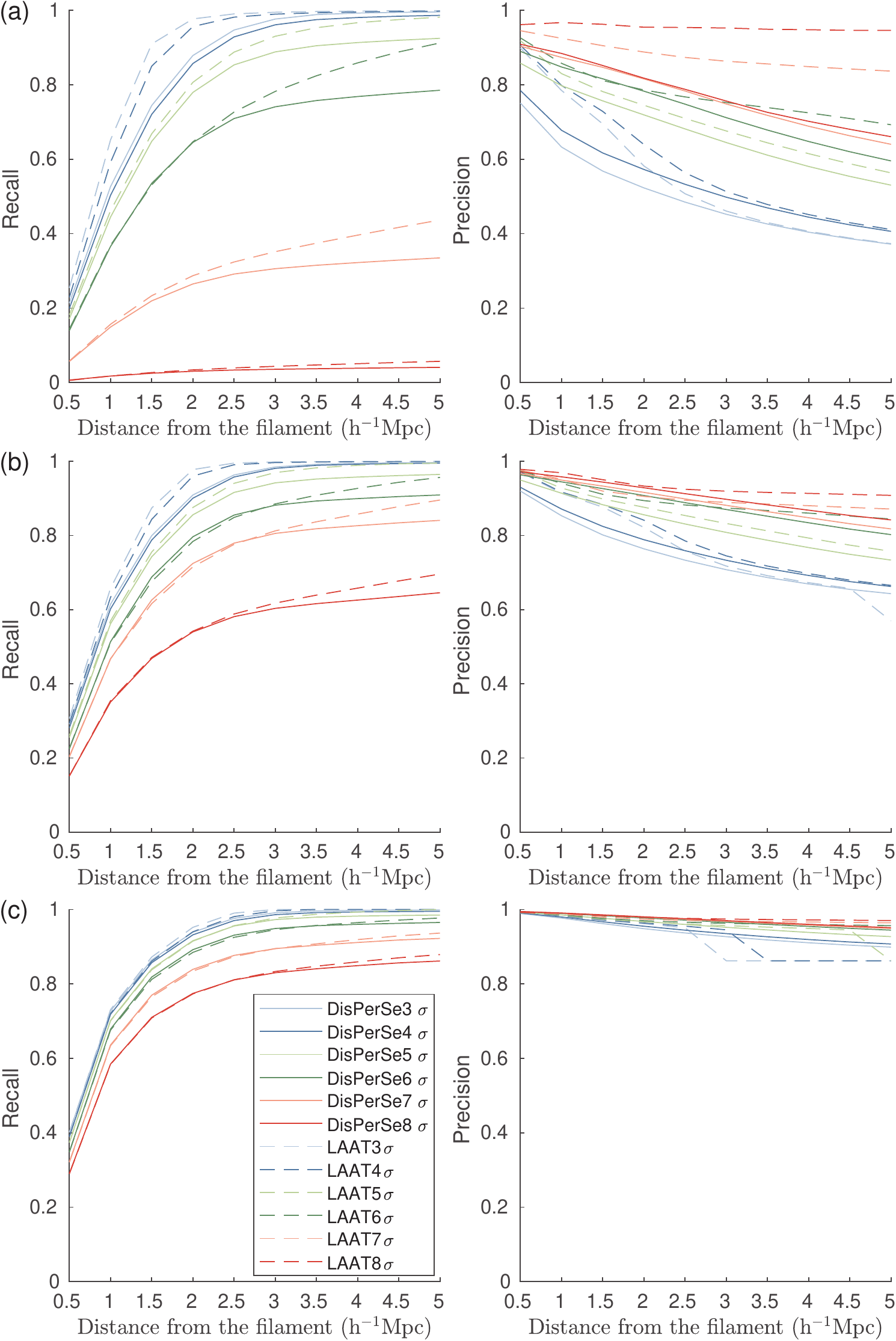}
    \caption{Precision and recall rates of the ant colony method with alignment versus Disperse for dataset A, B and C (a-c).
    }
    \label{PRcosmic}
\end{figure}

To compare the ant colony algorithm for manifold denoising with disperse we follow the same experiments as suggested in \cite{shivashankar_felix:_2016}, since the methods have different parameters. 
Disperse, for example, has a significance parameter (number of $\sigma$) which corresponds to the strength of the detected structure to the noise. 
Moreover, Disperse returns lines to resemble the found filaments and it does not convey any information about its thickness. 
Therefore, as suggested by \cite{shivashankar_felix:_2016}, we identify the points which are located within a fixed radius around the detected filament line as points on the filament. 
The solid lines in figure \ref{PRcosmic} present the performance in terms of Precision and Recall of Disperse for several significance values ($n\sigma$) and various radius sizes around the filament lines.

For the LAAT method, all parameters have been set to default values for all experiments, and the radius of neighborhood is set to $3$.
The only parameter which is altered is the threshold applied on the deposited pheromone on each point. We select the threshold value such that the number of remaining points is the same as the number of points after choosing a fixed strength value and fixed radius for Disperse. 
This process is iterated for several strength values and radii. The dashed lines in figure \ref{PRcosmic} present the results of LAAT on the three synthetic cosmic-web datasets with ground truth. Figure \ref{PRcosmic} illustrates as the number of positive labels increases from dataset A to C, the performance of both methods improves. 
Nonetheless, in most cases, with the same number of the survivor points as the output of Disperse and LAAT, both the precision and recall rates are higher for 
our method.
In Fig. \ref{PRcosmic} panel (c) as the number of false negatives goes to zero, the recall rate reaches one for $3\sigma$, $4\sigma$, and $5\sigma$ strength values. At this stage, the points which are labeled positive by Disperse include almost all the dataset points. However, since some of these points are seldom visited by ants, the amount of pheromone on many of them are almost the same and near to zero. Thus by applying a threshold on the result of LAAT, all the points in the dataset are selected, and the precision suddenly jumps to the lowest value possible.

\begin{figure}[t]
    \centering
    \includegraphics[width=\columnwidth]{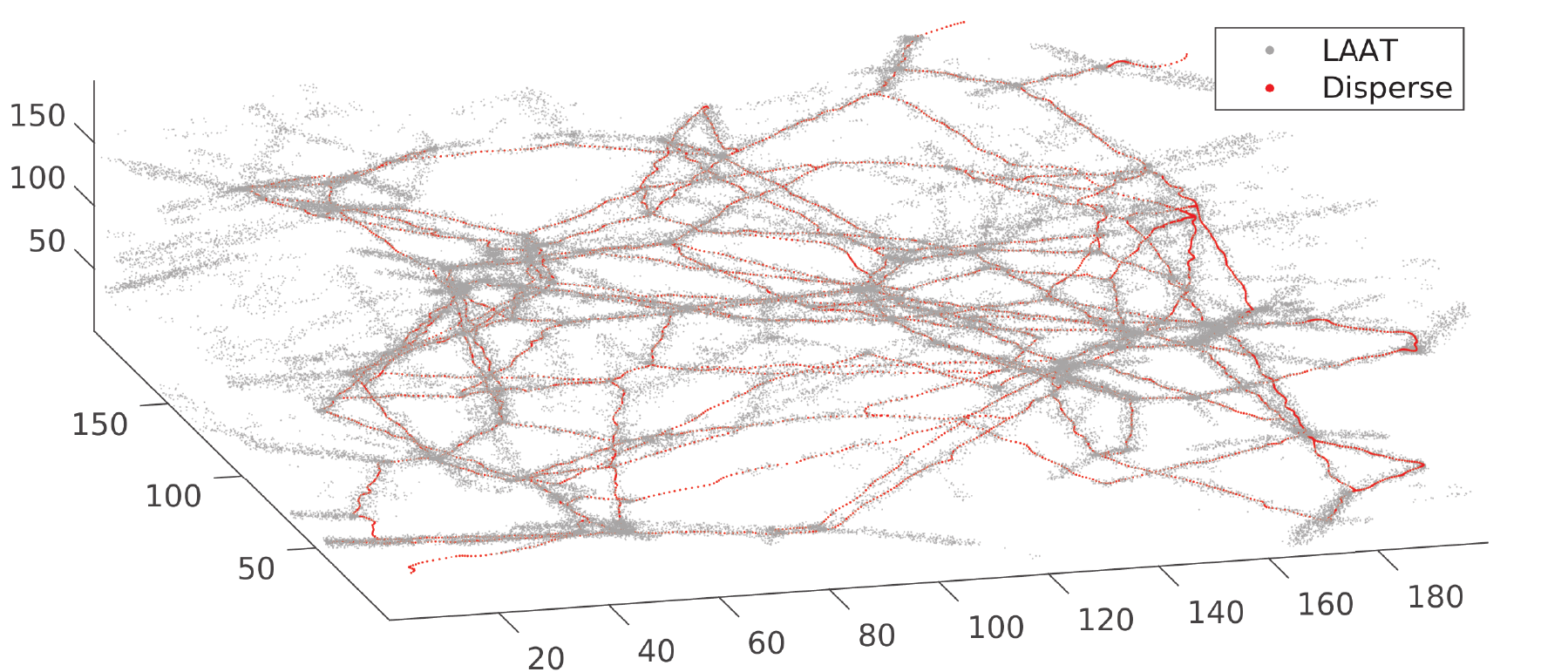}
    \caption{The results of LAAT and Disperse on dataset A}
    \label{LAATdisperse}
\end{figure}

Figure \ref{LAATdisperse} demonstrates the result of LAAT and Disperse applied to dataset A (see figure \ref{fig4}). 
For Disperse the strength value is fixed to $6\sigma$ and the resulting filament lines are visualized in red. 
The gray points denote the surviving points after applying the threshold to the LAAT result. 
Figure \ref{LAATdisperse} shows that Disperse did not reveal several even relatively strong filaments for this strength value. 
Moreover, it sometimes produces fake filaments in places where there are no actual filaments in the dataset, mostly near the borders. 
Although few points on the walls (labeled as noise) survive 
after applying the threshold, the LAAT successfully discovers most filaments with few noise points.

\subsection{Numerical comparison of LAAT, MC, and LLPD}
\label{compsection}
\begin{figure}[t]
    \centering
    \includegraphics[width=\columnwidth]{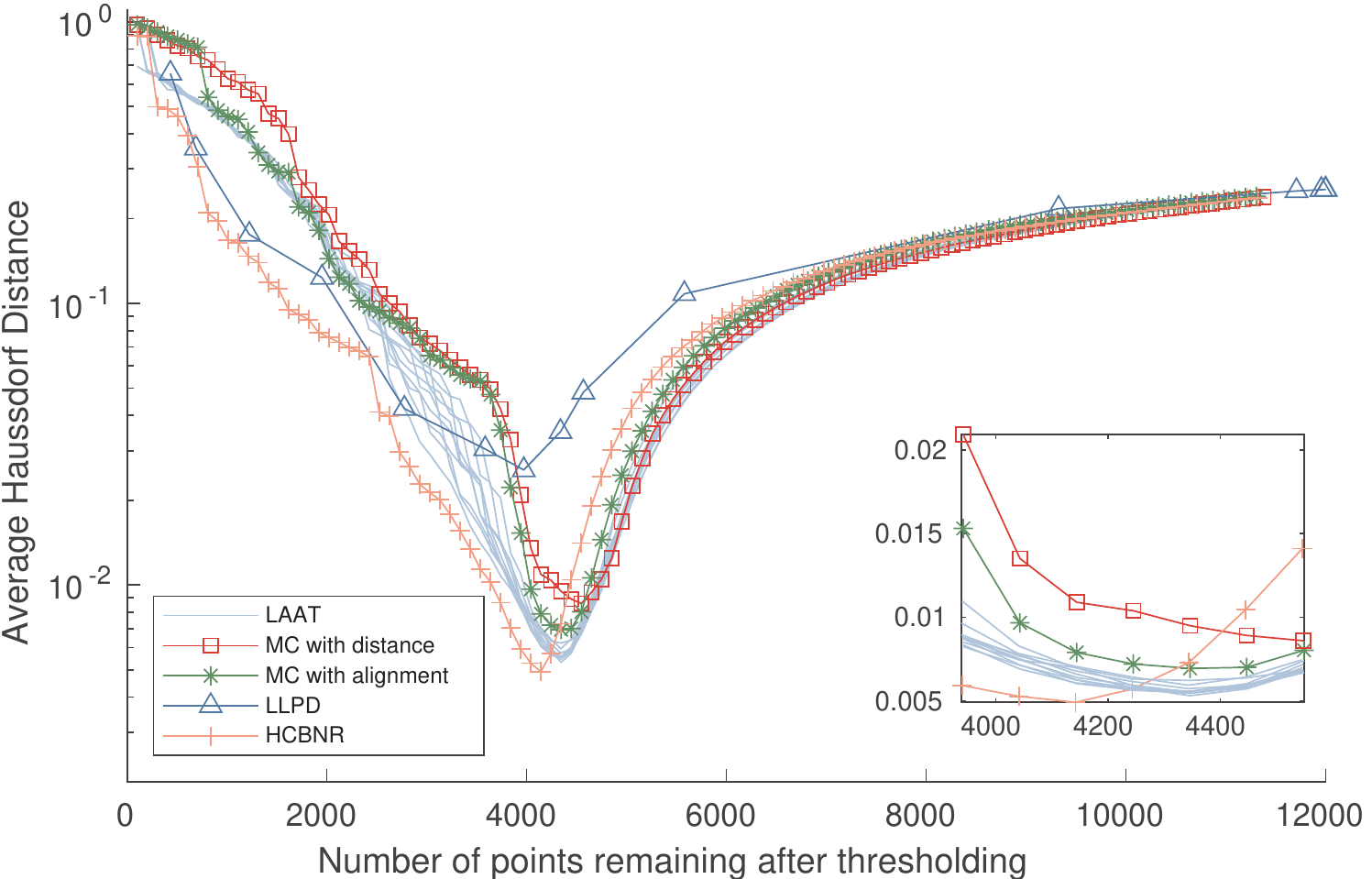}
    \caption{The AHD 
    for denoising methods on the two-arm dataset.
    The small box zooms the region 
    of the plot when the number of survived points is between 3900 to 4600.}
    \label{comp4method}
\end{figure}
To compare the LAAT, MC based on distance and alignment, and LLPD we start our investigation on the two-arm dataset.
All these methods use a threshold to distinguish noise points from the points on the manifolds and therefore we compare them over the complete range of possible threshold values.
The value was reduced from the maximum possible on a point to the minimum for each method, and the AHD between the surviving points after thresholding and the ground truth was computed. 
LAAT does not provide an analytical solution and thus we repeated the experiment 10 times and visualized the results of all three methods separately (see figure \ref{comp4method}).
Remarkably the AHD for LAAT is lower than MC based on alignment or distance information for most of the range. 
Interestingly, MC with a transition matrix based on alignment outperforms MC based on the distance between points, but only slightly. 
This indicates that the alignment information is useful for detecting manifold with curvature.

\begin{figure*}[t]
    \centering
    \includegraphics[width=2\columnwidth]{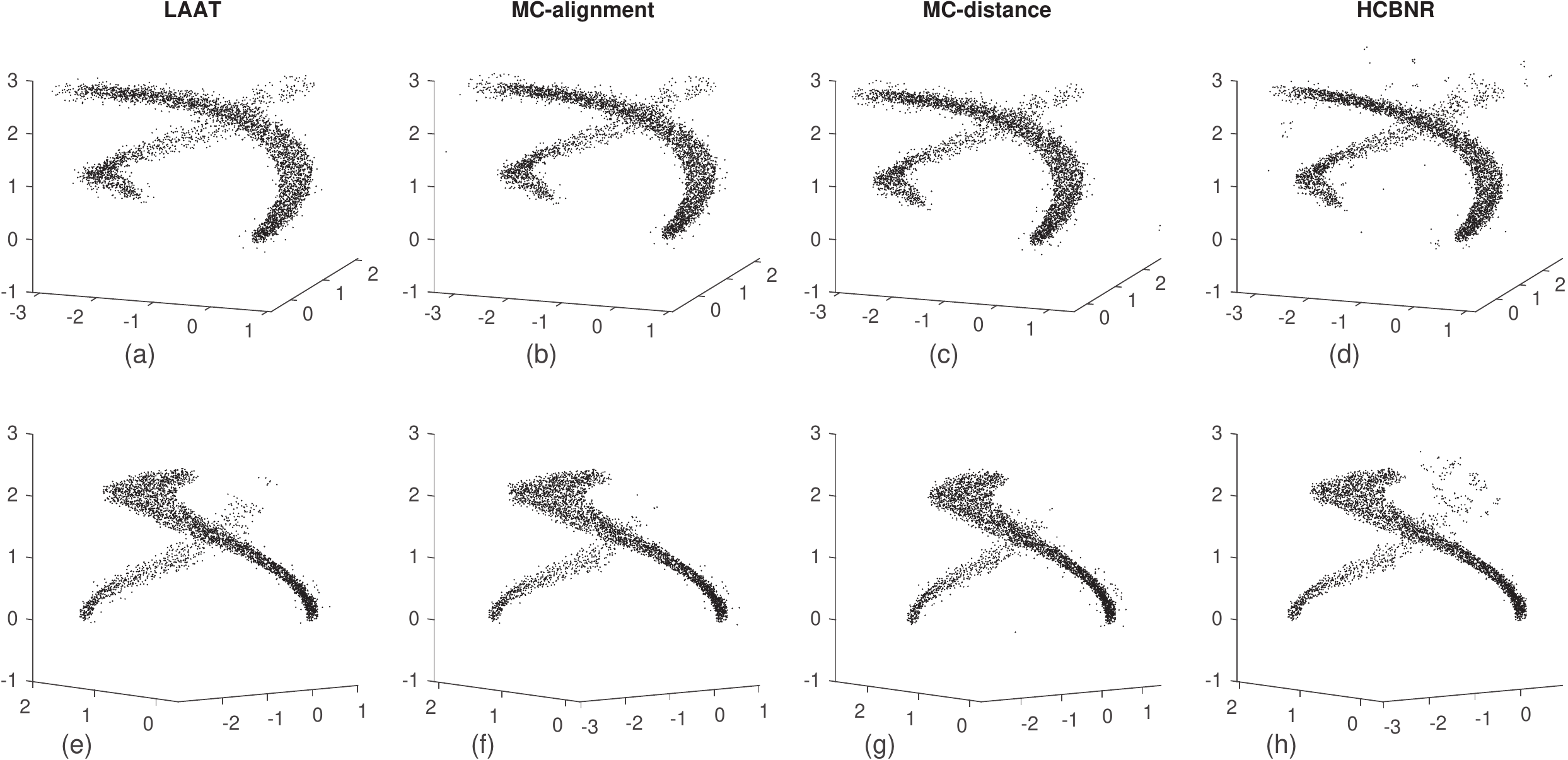}
    \caption{
    Examples of recovered points of the two-arm dataset with four compared methods. 
    The top row (a-d) shows 4500 recovered points (after the optimum indicated in the AHD plot Fig.~\ref{comp4method}) and the bottom row 3800 points (before the optimum).
    }
    \label{recover4method}
\end{figure*}

LLPD performs better each time it discovers the first points on a denser part of the arms until reaching roughly 3000 points. 
As more and more points are recovered with increasing thresholds, more and more noise points survive in comparison to other techniques, clearly shown by comparably high AHDs. 
ADBSCAN \cite{li_novel_2020} labels all detected noise points as one class, thus it is not plausible to examine its behavior over a range of thresholds. 
In contrast, HCBNR \cite{cheng_hierarchical_2019} 
uses the natural neighbor-based density and removes noise by imposing a threshold, 
which plays a similar role to that of the pheromone in LAAT. 
We observe that HCBNR 
generally recovers the two-arm dataset very well when the number of points requested is lower than the number of ``noiseless'' points on the manifold.
However, with increasing number of points HCBNR recovers more and more random noise points both close and far from the manifolds, which yields a sharper increase in AHD (roughly after 4000 points) in comparison with LAAT or MC. 
Figure \ref{recover4method} shows 3800 (top row) and 4500 (bottom row) points as recovered by each of the methods, namely Laat, MC with alignment, MC with Euclidean distance, and HCBNR.
The rows are plotted with two different viewpoints since the first row is designed to show the noise, while the second row 
demonstrates the detection of the low-density arm. 
As seen in the second row, HCBNR can discover the manifold points fairly well as long as they are denser than noise points, which is a very natural assumption.
Compared with HCBNR, LAAT recovers less of the second arm, but the MCs lose a big part of the low-density arm and include more noise. 
With an increased number of recovered points as shown in Figure \ref{recover4method}(a-d) the low-density arm is well recovered with all methods. 
However, the methods differ significantly in the recovered noise. 
HCBNR suddenly detects many dense noise clouds scattered in the background. MCs suffer from the same problem, albeit to a much smaller extent.
In contrast, the detected noise points by LAAT reside close to the manifold.

Table \ref{Tab} summarizes the AHD results on two synthetic datasets for the six methods. 
We determine the threshold for LLPD by obtaining the knee point in the plot of path distance to the 20th nearest neighbor as recommended in \cite{little_path-based_2020}. 
For ADBSCAN, the minimum number of samples is set to 20 as described in \cite{li_novel_2020}, and the noise percentage is determined based on prior knowledge on the synthetic datasets. 
Note that prior knowledge about the percentage of the noise is usually not available in real datasets, and hence this hyper-parameter might become difficult to tune.
For the other four methods, we select the threshold of the lowest AHD on the calibration dataset as is described before and apply the same threshold on the pheromone value 
on each point for LAAT, or on the vector which represents the steady-state situation of MC. 
The same strategy is used for the threshold on the natural neighbor-based density \cite{cheng_hierarchical_2019}.
Next, we compute the AHD between surviving points and ground truth. 
For LAAT the whole procedure was repeated ten times, and the number presents the mean value over all iterations. 
Table \ref{Tab} reveals that LAAT outperforms the other methods on both datasets as it depicts the lowest AHD compared to the rest. 
The AHD on the two-arm dataset for MC with a transition matrix based on alignment information is lower than the distance based MC. 
Nonetheless, the results are different on the four-cylinder dataset, which might suggest that the alignment information plays a crucial rule in discovering manifolds with curvature.
On the four-cylinders, the data points are spread uniformly in the volume, thus the jumping probabilities are similar in all directions. Hence, the performance of MC with alignment is similar to MC with distance.
\begin{table}[t]
\centering
\def\myCWidth{0.2}
\caption{AHD of methods for two-arm and four-cylinder.}
\label{Tab}
\begin{tabularx}{0.8\columnwidth}{ @{\extracolsep{\fill}}
l>{\raggedleft}p{\myCWidth\columnwidth}>{\raggedleft\arraybackslash}p{\myCWidth\columnwidth}
}
Method & \makecell[c]{Two-arm} & \makecell[c]{Four-cylinder} \\
\toprule
LAAT & $5.80\cdot10^{-3}$ & $1.42\cdot10^{-2}$ \\
\makecell[lt]{MC {\footnotesize(Alignment)}} & $6.96\cdot10^{-3}$ & $1.92\cdot10^{-2}$ \\
\makecell[lt]{MC {\footnotesize(Distance)}}  & $8.97\cdot10^{-3}$ & $1.79\cdot10^{-2}$ \\
LLPD    &$25.60\cdot10^{-3}$  & $5.69\cdot10^{-2}$ \\
HCBNR   & $7.13\cdot10^{-3}$  & $2.32\cdot10^{-2}$ \\
ADBSCAN & $7.43\cdot10^{-3}$  & $1.86\cdot10^{-2}$ \\
\bottomrule
\end{tabularx}
\end{table}

\subsection{
Comparison 
on N-body cosmological simulation}
\label{cosmicweb_sec}

\begin{figure}
    \centering
    \includegraphics[width=\columnwidth]{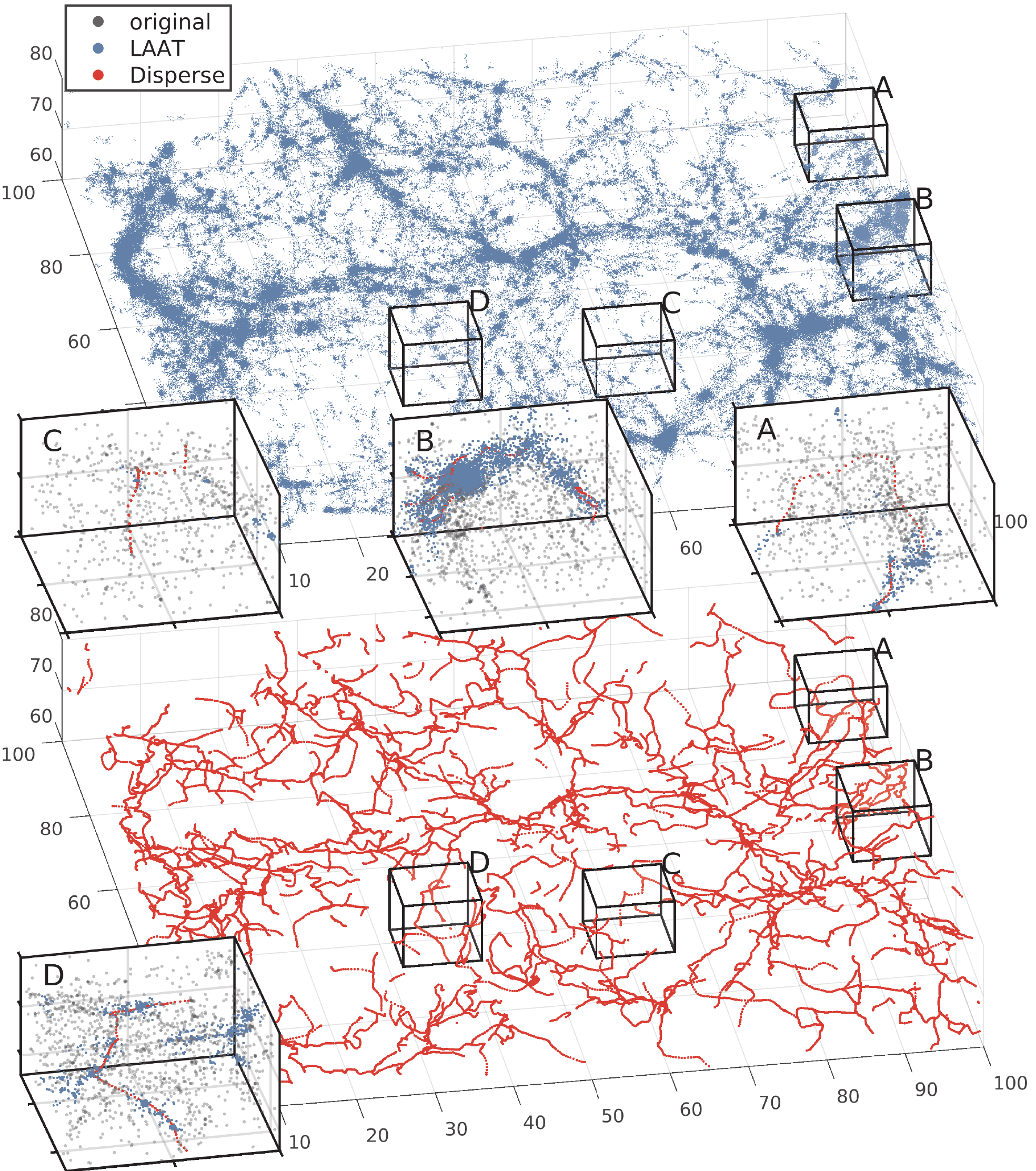}
    \caption{Results of LAAT and Disperse (top and bottom panel in blue and red) on the cosmic-web dataset. 
    Zoomed regions show examples of: 
    false positives introduced by Disperse (A and C), as well as 
    a strong filament found by LAAT but mostly missed by Disperse (B).
    Cube D shows a rare occasion where Disperse detects a filament more clearly than LAAT. 
    }
    \label{fig:zoom_web}
\end{figure}

We will now demonstrate
LAAT for detecting filaments in cosmological simulations of the cosmic web. 
In particular, we will compare LAAT with a widely used method in astrophysics, namely Disperse.
The methods are compared on the 6.7 million points real-world N-body simulation of the cosmic-web as illustrated in figure \ref{fig3}. 
For LAAT most parameters are kept at their previously mentioned default values while some are adapted to the data and application need. 
Specifically, we give more preference to the directional importance of the filaments and their one-dimensional nature by setting $\kappa=0.8$ and the neighborhood radius is set to 0.5. 
Due to the large number of samples and observing that the density notably varies on the manifolds we increase the number of ants and steps to make sure they cover all parts of the data ($N_\mathrm{ants}=200$, $N_\mathrm{steps}=12000$) and adapt the strategy of the ants placement in each epoch. 
Instead of placing ants on random points to start their walk we divide the whole data into 200 smaller cubes and in each iteration a single ant is placed randomly in each of them.
Due to the particularly steep density gradient in this dataset it is hard for the ants to escape the high-density regions, which is not avoided completely by increasing the number of ants and steps. 
Therefore, to catch the hugely varying densities we apply LAAT on the whole data, select all points the ants put pheromone on, and then exclude these points for a rerun on the remaining points. 
For this data and setting LAAT consumes about $8.7\mathrm{GB}$ of RAM, and both runs take less than 3 hours to compute the pheromone vector over a single core of a $4.5\mathrm{GHz}$ processor.
This can be reduced by parallelization as planned in future work. 

Structures are identified by both methods LAAT and Disperse by a thresholding strategy based on the Pheromone and strength parameter respectively. 
For LAAT our astronomical experts select an appropriate threshold for a small random cube covering 10\% of the volume of the whole dataset, which is then subsequently used for the full dataset. The strength parameter for Disperse is also set by experts to $7\sigma$. 
To make the filaments smoother each point is afterwards replaced by the mean of ten neighboring points within the same filament.
For this setup of the experiment, Disperse took 4.5 hours on a $3.5\mathrm{GHz}$ processor, and it occupies $80\mathrm{GB}$ of RAM. 
Note that Disperse run-time increases drastically if the sigma value is decreased to show fainter filaments.

The results of both methods applied to the same data cube can be found in figure \ref{fig:zoom_web}.
Disperse detects many filaments inside the data, but also identifies filaments in places in which there are no or few sample points. 
Increasing the strength parameter decreases the appearance of false ``ghost'' filaments at the cost of missing others. 
Notably, LAAT detects filaments of largely varying thickness along the filament. 
Altering the threshold value over a range on all points reveals weaker filaments and points which are located on walls (two-dimensional manifolds) with much lower density structure than the filaments.
The cubes A-D in figure \ref{fig:zoom_web} show zoomed regions that exemplify differences in the methods. Zoom B contains a quite strong and clear filament found by LAAT but disconnected and missed by Disperse. 
A and C show examples of ``ghost'' filaments introduced by Disperse in regions that do not appear to contain dense clear structures in the original data or marked by LAAT. 
Rarely one can find filaments that are nicely marked by Disperse and appear more disconnected by LAAT as depicted in zoom D. 
The qualitative comparison between the widely used Disperse and LAAT illustrates the versatility of our method capable of finding weak structures as well as filaments with strongly varying density robustly.

\section{Demonstration with more than 3 dimensions}
\label{high-dimension-section}
While Liebeskind et al. \cite{libeskind_tracing_2018} focus on large scale structures in 3D physical space, such as filaments of the cosmic web, astronomers are more and more frequently working with multi-dimensional data. 
An interesting higher dimensional problem to demonstrate the potential of LAAT is to help astronomers to detect star forming regions in galaxies. 
According to current 
theory 
star particles are created in regions of dense and cool gas and also high CII emission are associated with 
star formation \cite{DeLooze2011}. 
The CII emission rate (in~erg~s$^{-1}$~cm$^{-3}$) for each gas particle can be computed 
leveraging chemical evolution models of the galactic gas and taking into account its ionization equilibrium and ion level occupation \cite{Maio2007,DeRijcke2013}. 
%
Thus in the following, we demonstrate LAAT on an simulation snapshot of a 
galaxy containing 6 dimensions, including the three dimensions for the position of the gas particles, their density $\rho$, temperature $\tau$, and CII emission (photon emission at wavelength of 158 $\mu$m due to carbon ion C+).
Details on the physical models and simulation description used to obtain the galaxy snapshot under study, see \cite{Verbeke2017}. 

We take advantage of these extra dimensions and incorporate them as reward terms in equation \eqref{vformula} 
as follows:
\begin{equation}
V^{(i,j)}(t) = \kappa_1 \overline{F}^j(t)+\kappa_2 \overline{E}^{(i,j)}+\kappa_3 \overline{\rho}^{(i,j)} -\kappa_4 \overline{\tau}^{(i,j)} +\kappa_5 \overline{\text{CII}}^{(i,j)} .
\label{v6formula}
\end{equation}
Here, $\overline{\tau}^{(i,j)}$ and $\overline{\text{CII}}^{(i,j)}$ are defined analogously to 
\begin{equation}
\overline{\rho}^{(i,j)} = 
\sfrac{(\rho_j - \rho_i)}{      \sum\nolimits_{j^\prime\in \mathcal{N}^{(i)}_r} |\rho_{j'} - \rho_i|} \enspace.
\label{rhoformula}
\end{equation}
The sign of temperature in 
\eqref{v6formula} is negative, since the lower the temperature, the higher the chance of star formation.
\begin{figure*}[t]
    \centering
    \includegraphics[width=1.5\columnwidth]{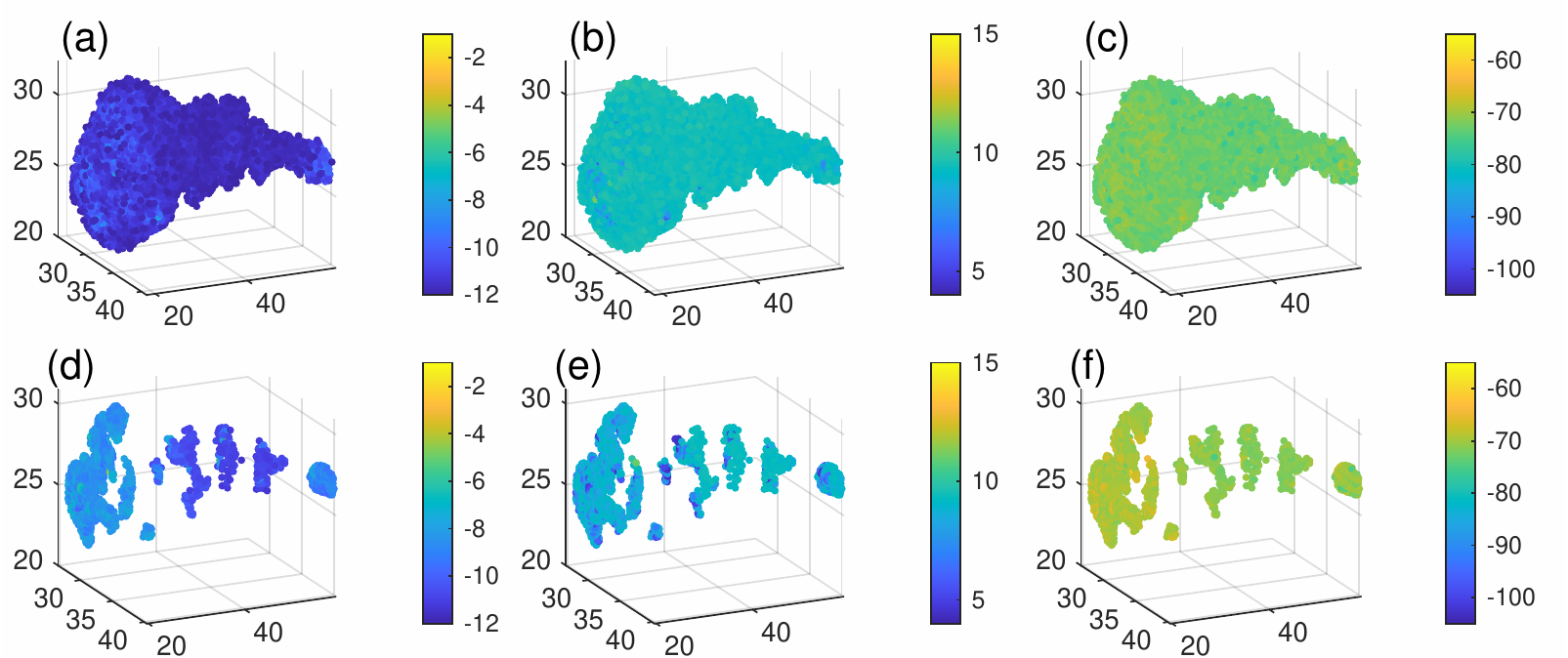}
    \caption{The color in a-d, b-e, and c-f pairs shows gas density, temperature, and CII in log scale, respectively. The second row of images(d, e, and f) illustrate the result of applying LAAT to highlight suitable conditions for star formation.}
    \label{6Dfigure}
\end{figure*}
Figure \ref{6Dfigure} depicts the simulated galaxy in 3D space color coded with the physical quantities $\rho$, $\tau$, and CII emission, respectively.
Here again, all parameters are set to default values except for $r=0.7$, $N_\mathrm{epoch}=20$, and $\kappa_c=0.2$ for all c in set $\{1,2,\dots,5\}$. The bottom row of figure \ref{6Dfigure} indicates that LAAT not only discloses points that belongs to manifolds but also make sure that these points satisfy the user-defined reward terms, namely high gas density, low temperature, and high CII emission which denote suitable conditions for star formation.

\section{Discussion on parameter setting}
\label{paramSet}
Generally, optimal parameter settings are data-dependent. 
They are influenced by the size, expansion, density, amount, and proportion of structures expected in the data, that the user is interested in. 
As demonstrated in the experiments the parameters are very robust, with a majority of values fixed despite the various data sets that differ greatly in size, density, and amount of structures. 
This section provides some discussion helpful for practical use.
The three variables $N_\mathrm{epoch}$, $N_\mathrm{ants}$, and $N_\mathrm{steps}$ determine the exploration power of the algorithm. 
As for MC, theoretically agents walk for an infinite number of steps, hence in practice we suggest to keep  $N_\mathrm{steps}$ much larger than the other two 
parameters\footnote{ 
Empirically we found that if the product of the three numbers is more than ten times the number of points in a dataset, LAAT converges.}.
The inverse temperature $\beta$ steers how much the system will randomize and hence its ability to explore the data space. 
As mentioned in section \ref{sec:AntColonyAlgo}, the remaining four parameters allow data and application specific tuning to structures the user is interested to find. 
The neighborhood radius $r$ determines the level of curvature and level of faintness that can be detected. 
It is a parameter that is also found in MCs, but we often observe that LAAT is more robust with it (see section \ref{phBenefit}).  
In general, $r$ should be large enough for local PCA to capture the main components of the local structure, nevertheless, it also should not be too large since it 
increases the computational cost. 
The pheromone influences the reinforcement and is determined with the absolute amount deposited ($\varphi$) and 
the evaporation rate ($\zeta$). 
Their interplay determines the size and the part of structures detected dependent on their density. 
For many structures with a largely varying density, a generous deposition and slow evaporation are advised. 
Lastly $\kappa\in [0,1]$ allows to tune between the alignment ($\kappa\rightarrow1$) and pheromone ($\kappa\rightarrow0$) preference. 
Hence, if mostly filament-like structures are expected, as for the cosmic web in section \ref{cosmicweb_sec}, 
we can give more attention to directional information by setting $\kappa$ close to 1.
Generally, our experiments in section \ref{paramset} for the two-arms and four-cylinders data show very robust results against changing $\beta$, $\kappa$, $\varphi$, $\zeta$ and hence we defined a set of default parameters. 
In all conducted experiments, we set these four parameters in the same range as discussed.

\section{Conclusion}
\label{sec:conclusions}
We propose the Locally Aligned Ant Technique (LAAT) to detect multiple manifolds buried in a large number of noise and outliers.
The method is inspired by the heuristic ant colony algorithm extended by the incorporation of local alignment information and a reinforcing pheromone term. 
Section \ref{CompMC} details the relationship to the Markov Chain, which can be considered a special case of LAAT when no pheromone is released.
We quantitatively demonstrated and compared the performance of alternative techniques on several synthetic data sets of varying complexity evaluated empirically using the 
AHD to the underlying noiseless manifolds. 
The experiments reveal that LAAT performs similarly or superior to alternative methods, 
robustly detecting structures of high curvature, varying density and even finding multiple manifolds with distinct features simultaneously.
We illustrate our method on two large real-world astronomical datasets, a 6.7 million point N-body simulation of the cosmic-web which is qualitatively compared to a 
technique widely used by astronomical experts. 
Unlike \cite{sousbie_Disperse_2011-1,sousbie_Disperse_2011_2}, and \cite{shivashankar_felix:_2016} our method can naturally deal with more than three dimensions, 
and we demonstrate how experts can use extra information in the form of reward terms to prescribe properties of structures to detect.

\vspace*{-0.3cm}
\section*{Acknowledgment}
\footnotesize
\begin{wrapfigure}[4]{L}{0.17\columnwidth} 
\vspace*{-0.35cm}
    \includegraphics[width=0.17\columnwidth]{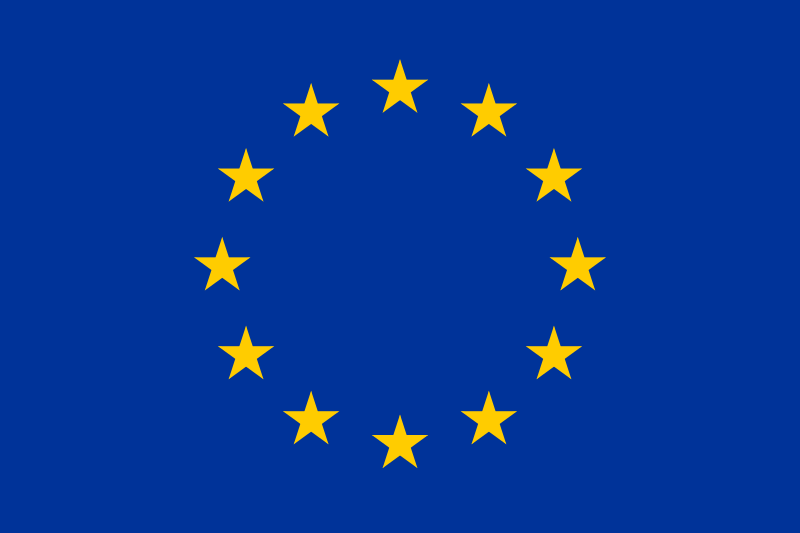} 
\end{wrapfigure}
\noindent 
This project has received financial support from the European Union's Horizon 
2020 research and innovation program under the Marie Sk{\l}odowska-Curie 
grant agreement No. 721463 to the SUNDIAL ITN network.
%
We furthermore 
thank the Center for Information Technology of the University of Groningen providing 
access to the Peregrine high performance 
cluster.
\ifCLASSOPTIONcaptionsoff
  \newpage
\fi


%

\bibliography{ms}
\vspace{-1.5 cm}
\bibliographystyle{ieeetr}
\def\myWidth{1in}\def\myHeight{1.25in}
\begin{IEEEbiography}[{\includegraphics[width=\myWidth,height=\myHeight,clip,keepaspectratio]{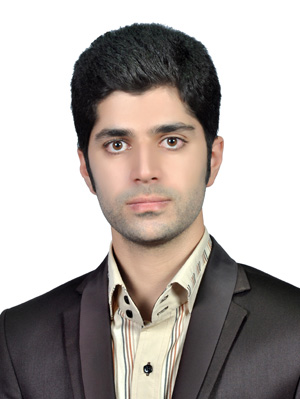}}]{Abolfazl Taghribi}
is a PhD student at the Faculty of science and engineering of the University of Groningen in The Netherlands. He received his bachelor and master degrees from Amirkabir University of Technology in Electronic engineering. His research interests include machine learning, algebraic topology, and computer vision.
\end{IEEEbiography}
\vskip -2\baselineskip plus -1fil
\begin{IEEEbiography}[{\includegraphics[width=\myWidth,height=\myHeight,clip,keepaspectratio]{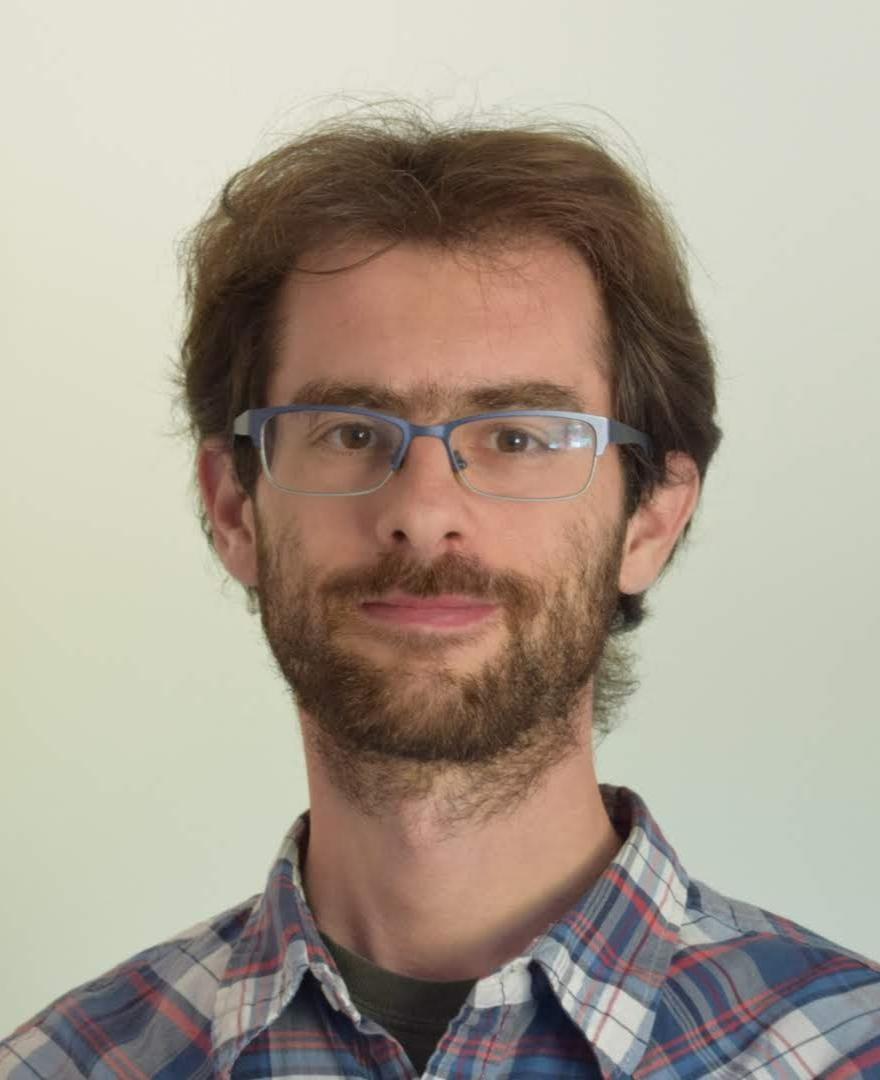}}]{Michele Mastropietro}
is a PhD student in Physics and Astronomy at the University of Ghent, Belgium. He received his bachelor and master degrees in Aerospace Engineering from University of Pisa, Italy. His research interests include astronomy, numerical simulations, machine learning, data visualization.
\end{IEEEbiography}

\vskip -2\baselineskip plus -1fil

\begin{IEEEbiography}[{\includegraphics[width=\myWidth,height=\myHeight,clip,keepaspectratio]{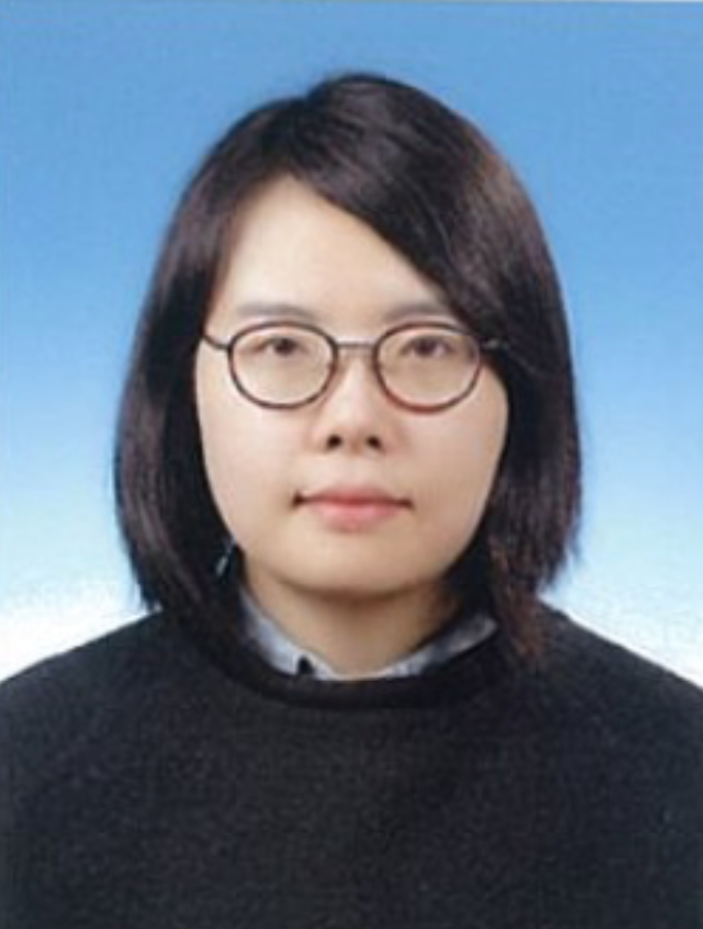}}]{Jihye Shin} (Ph.D from Kyung Hee University, S. Korea) obtained a KIAA fellowship at Kavli Institute for Astronomy and Astrophysics, Peking University, China. In 2016, she moved to Korea Institute for Advanced Study. Since 2018, she is currently a staff research scientist at Korea Astronomy and Space science Institute. Her current research interests are related to massive numerical simulations to study galaxy formation, and cosmological large-scale structures.
\end{IEEEbiography}

\vskip -2\baselineskip plus -1fil

\begin{IEEEbiography}[{\includegraphics[width=\myWidth,height=\myHeight,clip,keepaspectratio]{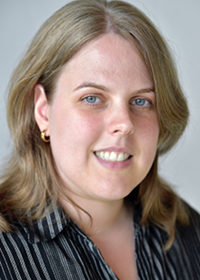}}]{Kerstin Bunte} 
is an assistant professor and Rosalind Franklin Fellow at the Bernoulli Institute for Mathematics, Computer Science and Artificial Intelligence (University of Groningen, The Netherlands) since July 2016. 
In 2015 she got a European Marie Sk{\l}odowska-Curie Fellowship (Project ID: 659104) at the University of Birmingham and is partner in the European ITN: SUNDIAL, Project ID: 721463. 
Her recent work has focused on the development of interpretable machine learning techniques for interdisciplinary data analysis and visualization. 
Further information can be obtained from http://www.cs.rug.nl/$\sim$kbunte/.
\end{IEEEbiography}

\vskip -2\baselineskip plus -1fil

\begin{IEEEbiography}[{\includegraphics[width=\myWidth,height=\myHeight,clip,keepaspectratio]{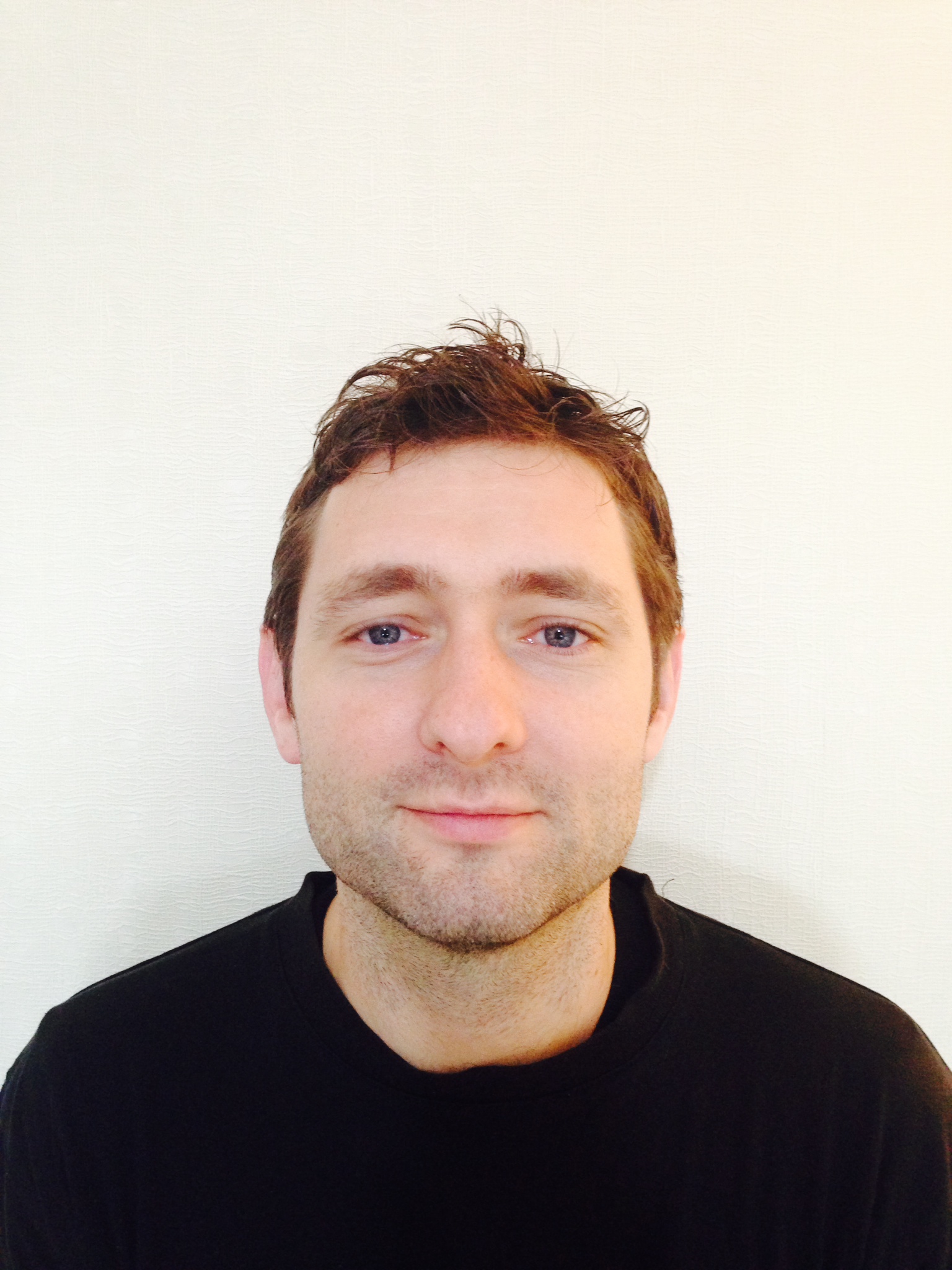}}]{Rory Smith} (Mphys and Ph.D from University of Wales, Cardiff) was a lecturer at Cardiff University. He spent 5 years as a researcher at Concepción university and was awarded a FONDECYT fellowship. He then moved to Yonsei University in Seoul, S. Korea for a research professorship. He was a staff research scientist at KASI (the Korean Astronomy and Space science Institute) in Daejeon, S. Korea for four years within the Galaxy evolution group. He recently returned to Chile as a lecturer and researcher at the Universidad Técnica Federico Santa Maria. His
interests are related to using numerical simulations to study galaxies and the impact of interactions with each other and their environment on their evolution. For more information, please see: https://rorysmith274.wixsite.com/mysite
\end{IEEEbiography}

\vskip -2\baselineskip plus -1fil

\begin{IEEEbiography}[{\includegraphics[width=\myWidth,height=\myHeight,clip,keepaspectratio]{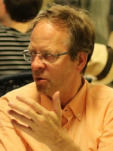}}]{Reynier F. Peletier} (M.Sc. University of Leiden. Ph.D. University of Groningen) 
is currently a Full Professor at the Kapteyn Astronomical Institute at the University of Groningen. He is interested in the evolution of galaxies, their dynamics and stellar populations, but also in instrumentation for large astronomical telescopes. He has edited several books and written more than 200 papers in refereed journals. At present, he is leading an EU International Training Network, named SUNDIAL, integrating computer scientists and astronomers to develop novel data analysis techniques for use in astronomy.
\end{IEEEbiography}

\vskip -2\baselineskip plus -1fil

\begin{IEEEbiography}[{\includegraphics[width=\myWidth,height=\myHeight,clip,keepaspectratio]{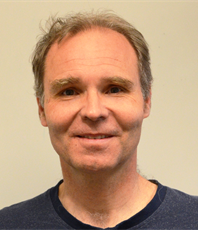}}]{Peter Ti\v{n}o}
(M.Sc. Slovak University of Technology, Ph.D. Slovak Academy of Sciences) 
is with the School of Computer Science, University of Birmingham, 
where he is currently a Full Professor-Chair in Complex and Adaptive Systems. His current research interests include dynamical systems, machine learning, probabilistic modelling of structured data, evolutionary computation, and fractal analysis. Peter was a recipient of the U.K.- Hong-Kong Fellowship for Excellence in 2008, three Outstanding Paper of the Year Awards from the IEEE Trans. on Neural Networks in 1998 and 2011 and the IEEE Trans. on Evolutionary Computation in 2010, and the Best Paper Award at ICANN 2002. 
\end{IEEEbiography}




\end{document}